\documentclass[sigconf]{acmart}

\AtBeginDocument{%
  }

\copyrightyear{2026}
\acmYear{2026}
\setcopyright{cc}
\setcctype{by}
\acmConference[WSDM '26]{Proceedings of the Nineteenth ACM International Conference on Web Search and Data Mining}{February 22--26, 2026}{Boise, ID, USA}
\acmBooktitle{Proceedings of the Nineteenth ACM International Conference on Web Search and Data Mining (WSDM '26), February 22--26, 2026, Boise, ID, USA}
\acmPrice{}
\acmDOI{10.1145/3773966.3777971}
\acmISBN{979-8-4007-2292-9/2026/02}

\usepackage{bm}
\usepackage{algorithm}
\usepackage{algpseudocode}
\usepackage{hyperref}
\usepackage{microtype}
\usepackage{mdframed}
\usepackage{multirow}
\usepackage{makecell}
\acmSubmissionID{490}

\usepackage{amsmath}
\usepackage{graphicx} 
\usepackage{balance}

\usepackage{pifont}   
\newcommand{\cmark}{\ding{51}} 
\newcommand{\xmark}{\ding{55}} 
\begin{document}
\begin{sloppypar}
\title{PR-CapsNet: Pseudo-Riemannian Capsule Network with Adaptive Curvature Routing for Graph Learning}
\author{Ye Qin}
\orcid{0009-0002-7098-3041}
\affiliation{%
  \institution{Guangdong University of Technology}
  \city{Guangzhou}
  \country{China}
}
\email{3122000617@mail2.gdut.edu.cn}

\author{Jingchao Wang}
\orcid{0000-0002-0099-539X}
\affiliation{%
  \institution{Peking University}
  \city{Beijing}
  \country{China}
}
\email{ethanwangjc@163.com}

\author{Yang Shi}
\orcid{0009-0009-3928-7495}
\affiliation{%
  \institution{Guangdong University of Technology}
  \city{Guangzhou}
  \country{China}
}
\email{sudo.shiyang@gmail.com}

\author{Haiying Huang}
\orcid{0009-0001-3303-3111}
\affiliation{%
  \institution{Guangdong University of Technology}
  \city{Guangzhou}
  \country{China}
}
\email{3222004766@mail2.gdut.edu.cn}

\author{Junxu Li}
\orcid{0009-0008-2846-9931}
\affiliation{%
  \institution{Guangdong University of Technology}
  \city{Guangzhou}
  \country{China}
}
\email{3120004701@mail2.gdut.edu.cn}

\author{Weijian Liu}
\orcid{0009-0002-5470-9878}
\affiliation{%
  \institution{Guangdong University of Technology}
  \city{Guangzhou}
  \country{China}
}
\email{1229523824@qq.com}

\author{Tinghui Chen}
\orcid{0009-0007-9849-9338}
\affiliation{%
  \institution{Guangdong University of Technology}
  \city{Guangzhou}
  \country{China}
}
\email{3122002219@mail2.gdut.edu.cn}

\author{Jinghui Qin}
\authornote{Corresponding author.}
\orcid{0000-0003-0663-199X}
\affiliation{%
  \institution{Guangdong University of Technology}
  \city{Guangzhou}
  \country{China}
}
\email{qinjinghui@gdut.edu.cn}

\renewcommand{\shortauthors}{Ye Qin et al.}
\begin{abstract}
Capsule Networks (CapsNets) show exceptional graph representation capacity via dynamic routing and vectorized hierarchical representations, but they model the complex geometries of real-world graphs (e.g., hierarchies, clusters, cycles) poorly by fixed-curvature space due to the inherent geodesical disconnectedness issues, leading to suboptimal performance. 
Recent works find that non-Euclidean pseudo-Riemannian manifolds provide specific inductive biases for embedding graph data, but how to leverage them to improve CapsNets is still underexplored. Here, we extend the Euclidean capsule routing into geodesically disconnected pseudo-Riemannian manifolds and derive a \textbf{P}seudo-\textbf{R}iemannian \textbf{Caps}ule \textbf{Net}work (\textbf{PR-CapsNet}), which models data in pseudo-Riemannian manifolds of adaptive curvature, for graph representation learning. Specifically, PR-CapsNet enhances the CapsNet with Adaptive Pseudo-Riemannian Tangent Space Routing by utilizing pseudo-Riemannian geometry.
Unlike single-curvature or subspace-partitioning methods, PR-CapsNet concurrently models hierarchical and cluster/cyclic graph structures via its versatile pseudo-Riemannian metric. It first deploys \textbf{Pseudo-Riemannian Tangent Space Routing} to decompose capsule states into spherical-temporal and Euclidean-spatial subspaces with diffeomorphic transformations. Then, an \textbf{Adaptive Curvature Routing} is developed to adaptively fuse features from different curvature spaces for complex graphs via a learnable curvature tensor with geometric attention from local manifold properties. 
Finally, a geometric properties-preserved \textbf{Pseudo-Riemannian Capsule Classifier} is developed to project capsule embeddings to tangent spaces and use curvature-weighted softmax for classification.
Extensive experiments on node and graph classification benchmarks show PR-CapsNet outperforms state-of-the-art models, validating PR-CapsNet’s strong representation power for complex graph structures.
\end{abstract}




\maketitle

\section{Introduction}
Learning on graph-structured data is crucial yet significantly challenging for real-world applications due to the inherent complexity of intricate relationships among nodes. As a solution, Graph Representation Learning (GRL) aims to generate graph representation vectors that capture the graph structure and node features of large graphs accurately. Mainstream Graph Neural Networks (GNNs)~\cite{26, velickovic2018graph, wu2019simplifying, bruna2014spectral} have demonstrated success through an iterative local aggregation mechanism within GRL. However, they mainly rely on simple linear aggregation in Euclidean space. Even with recent advancements like Graph Transformers~\cite{58}, which enhance global modeling, they often risk overfitting on node classification tasks and lack specific geometric inductive biases. Consequently, these Euclidean-based methods are limited in capturing intricate structural patterns and fine-grained relationships in complex graphs~\cite{boguna2021network, gu2019learning}.

To address the limitations of scalar-based aggregation in GNNs, Capsule Network (CapsNet)~\cite{45} offers a promising alternative by utilizing vectorized representations (capsules) and dynamic routing. Unlike GNNs that may suffer from over-smoothing, CapsNets can disentangle complex part-whole relationships and preserve richer structural information and spatial relationships for capturing accurate structural patterns. 
Recent works~\cite{28, 29} have extended CapsNets to graph data for more selective and perceptive aggregation in Euclidean space. Recognizing graph structures' non-Euclidean character, which is incompatible with Euclidean space, researchers made attempts to explore non-Euclidean geometries to improve the capacities of CapsNets on graph data. Hyperbolic Capsule Networks~\cite{41, 42} explore hyperbolic space with negative curvature, which excels at embedding graphs with strong hierarchical structures~\cite{krioukov2010hyperbolic, 6, boguna2021network}. Meanwhile, some works~\cite{wilson2014spherical, meng2019spherical, defferrard2019deepsphere, davidson2018hyperspherical} explore spherical spaces more suitable for graphs with cyclic or clustered structures. 
		
Real-world graphs are complex and often exhibit different geometric paradigms, such as the combinations of hierarchical and clustered components~\cite{boguna2021network, gu2019learning}. These graphs can not be handled well in Euclidean geometry and should be processed in non-Euclidean space. However, fixed-curvature Riemannian methods~\cite{gu2019learning, laub2004feature} struggle with varying local geometric properties, leading to mismatched representations. Those methods combining multiple Riemannian spaces or using product manifolds also face critical  limitations~\cite{boguna2021network, gu2019learning}, including strict subspace independence with limited interactions between geometric components, complex cross-subspace alignment, and the inability to capture the coexistence of positive and negative curvature within the same graph structure. Pseudo-Riemannian geometry is a promising theory that can address these limitations by allowing indefinite metric tensors with positive and negative eigenvalues. It can model mixed geometric features within a single manifold uniformly with well-defined geometric operations while avoiding explicit subspace partitioning~\cite{oneill1983semi, 15, 13}, but how to extend CapsNets with pseudo-Riemannian geometry for complex graphs is challenging and underexplored.

In this paper, we extend the Euclidean capsule routing into geodesically disconnected pseudo-Riemannian manifolds and derive a \textbf{P}seudo-\textbf{R}iemannian \textbf{Caps}ule \textbf{Net}work (\textbf{PR-CapsNet}) that models data in pseudo-Riemannian manifolds of adaptive curvature in the context of graph representation learning. Specifically, PR-CapsNet enhances the Capsule Network with Adaptive Pseudo-Riemannian Tangent Space Routing by utilizing pseudo-Riemannian geometry.
Unlike single-curvature or subspace-partitioning methods, PR-CapsNet concurrently models hierarchical (negative curvature) and cluster/cyclic (positive curvature) graph structures via its versatile pseudo-Riemannian metric. It first deploys \textbf{Pseudo-Riemannian Tangent Space Routing} to decompose capsule states into spherical-temporal and Euclidean-spatial subspaces using diffeomorphic transformations, resolving geodesical disconnectedness issues\footnote{In this paper, 'temporal' and 'spatial' refer purely to the pseudo-Riemannian metric signature components, rather than any real-time or dynamic behaviour.}. Then, an \textbf{Adaptive Curvature Routing} is developed to adaptively fuse features from different curvature spaces for complex graphs via a learnable curvature tensor with geometric attention from local manifold properties. 
Finally, a \textbf{Pseudo-Riemannian Capsule Classifier} is developed to project capsule embeddings to tangent spaces and use curvature-weighted softmax for classification while preserving geometric properties. 
Extensive experiments on node and graph classification benchmarks show PR-CapsNet outperforms state-of-the-art models, validating our pseudo-Riemannian framework's enhancement of CapsNets’ representational power for complex graph structures.

Overall, our main contributions can be summarized as follows. First, we propose a Pseudo-Riemannian Capsule Network (PR-CapsNet) with a novel dynamic routing mechanism to model data in Pseudo-Riemannian manifolds of adaptive curvature in the context of graph representation learning.
Second, we introduce an Adaptive Curvature Routing (ACR) module based on Pseudo-Riemannian Tangent Space Routing that allows PR-CapsNet to dynamically learn and leverage features processed through different geometric perspectives, enhancing its ability to model graphs with diverse and varying local geometric characteristics.
Third, we develop a geometric properties-preserved Pseudo-Riemannian Capsule Classifier (PRCC) to project capsule embeddings to tangent spaces and use curvature-weighted softmax for classification. 
Finally, extensive experiments on node classification and graph classification benchmarks demonstrate that our PR-CapsNet framework achieves state-of-the-art performance, validating the effectiveness of leveraging pseudo-Riemannian geometry within a CapsNet architecture for complex graph structures.

\section{Related Work}
\subsection{Pseudo-Riemannian Manifolds}
Traditional data is often converted into embeddings in Euclidean space~\cite{1,2} due to its simple geometric properties. However, Euclidean geometry distorts tree-like structures due to mismatched growth patterns~\cite{3}, and many complex data types (e.g., graphs) exhibit non-Euclidean properties~\cite{4}. Hyperbolic space~\cite{5} can better preserve the topology of hierarchical structures through its curved geometry, showing advantages in representing such data~\cite{6,7}. Those GCNs based on hyperbolic space achieve impressive performance~\cite{8,9,10,11,12}. However, the constant negative curvature of hyperbolic space limits its ability to model complex graphs with mixed topologies~\cite{13}. Pseudo-Riemannian manifolds with non-zero curvature were also explored for graph embeddings~\cite{14,15,16}, but initially lacked adequate geodesic tools. Pseudo-Riemannian GCN~\cite{13} advanced this by extending GCNs to these manifolds with introducing novel geodesic tools and diffeomorphic operations to handle geodesic disconnection, enabling more flexible graph representations.

\subsection{CapsNets in Graph Learning}
Convolutional Neural Networks (CNNs)~\cite{17,18,19}, strong feature extractors for computer vision, struggle with spatial hierarchies and viewpoint variations. Graph Neural Networks (GNNs)~\cite{20,21,22,23} were introduced for complex data relationships. Graph Convolutional Networks (GCNs)~\cite{24,25,26} learn graph representations by aggregating information, but struggle with hierarchical structures and complex dependencies.
Enhancing GNNs with Capsule Networks (CapsNets)~\cite{27,28,29} can improve robustness. CapsNets~\cite{45} preserve spatial feature relationships and improve transformation robustness, succeeding in NLP~\cite{30,31,32} and medical imaging~\cite{33,34,35}. Routing algorithms~\cite{36,37,38} and dynamic routing mechanisms~\cite{39,40} further enhance CapsNets' performance. More recently, Hyperbolic Capsule Networks~\cite{41,42} integrate CapsNets with hyperbolic embeddings to enhance hierarchical representation learning, effectively capturing complex structures. 
Although hyperbolic embeddings can capture tree-like structures~\cite{6}, their fixed negative curvature limits modeling complex graphs with mixed topologies~\cite{3}. CapsNets excel at part-whole relationships~\cite{45} but face challenges in non-Euclidean settings. Extending them to Pseudo-Riemannian spaces requires addressing geodesic completeness, feature consistency across varying curvatures, and adapting Euclidean routing mechanisms to curvature-adaptive embeddings to avoid geometric inconsistency.

\section{Preliminary}
\label{sec:preliminary}

\subsection{Pseudo-Riemannian Manifolds}
\label{subsec:pseudo_riemannian}

Pseudo-Riemannian geometry generalizes Riemannian geometry by allowing an indefinite metric tensor, which can yield positive, negative, or zero squared lengths for tangent vectors. This property enables the construction of manifolds with rich and varied geometric structures. 
\begin{definition}[Pseudo-Euclidean Space]
A Pseudo-Euclidean space $\mathbb{R}^{s,t+1}$ is an $(s+t+1)$-dimensional vector space equipped with a non-degenerate, symmetric bilinear form. The Pseudo-Euclidean inner product $\langle \cdot, \cdot \rangle_{\text{ps}}$ for $\mathbf{x}, \mathbf{y} \in \mathbb{R}^{s,t+1}$ is defined as follows:
\begin{equation}
    \langle \mathbf{x}, \mathbf{y} \rangle_{\text{ps}} = -\sum_{k=1}^{s} \mathbf{x}_k \mathbf{y}_k + \sum_{k=s+1}^{s+t+1} \mathbf{x}_k \mathbf{y}_k,
    \label{eq:pseudo_inner_product}
\end{equation}
where the first $s$ dimensions are often referred to as "space-like" and the remaining $t+1$ dimensions as "time-like". The signature of the metric is $(s, t+1)$.
\end{definition}
Within this ambient space, specific manifolds can be defined. Our focus is on the pseudo-hyperboloid, which is a manifold of constant non-zero curvature.
\begin{definition}[Pseudo-hyperboloid $\mathcal{Q}_{s,t}^\beta$]
The pseudo-hyperboloid $\mathcal{Q}_{s,t}^\beta \subset \mathbb{R}^{s,t+1}$ is a set of points satisfying the following condition:
\begin{equation}
    \mathcal{Q}_{s,t}^\beta = \{ \mathbf{x} \in \mathbb{R}^{s,t+1} : \langle \mathbf{x}, \mathbf{x} \rangle_{\text{ps}} = \beta \},
    \label{eq:pseudo_hyperboloid_def}
\end{equation}
where $\beta \in \mathbb{R} \setminus \{0\}$ is a constant determining the curvature. If $\beta < 0$, it is a pseudo-hyperboloid of one sheet or two sheets, depending on conventions and specific components if $s=0$. If $\beta > 0$, it is a pseudo-sphere.
\end{definition}
A significant challenge when working with pseudo-hyperboloids is their property of being \textbf{\emph{geodesically disconnected}}~\cite{13} that not all pairs of points on the manifold can be connected by a geodesic path. Consequently, standard geometric tools like the logarithmic map, which maps a point on the manifold to a tangent vector at another point, are not globally well-defined. This issue is particularly intractable for operations like aggregation or distance computation, which are fundamental in many learning algorithms.

To overcome this, Pseudo-Riemannian GCN~\cite{13} introduces a set of \emph{diffeomorphic geodesic tools} built upon a crucial diffeomorphism that maps the pseudo-hyperboloid to a geodesically connected product space.
\begin{theorem}[Diffeomorphism $\psi$]
There exists a diffeomorphism $\psi: \mathcal{Q}_{s,t}^\beta \to S_t^{-\beta} \times \mathbb{R}^s$, where $S_t^{-\beta}$ is a $t$-dimensional sphere of radius $\sqrt{|-\beta|}$ (if $\beta < 0$) or a hyperbolic space (if $\beta > 0$ and $S_t^{-\beta}$ is interpreted accordingly), and $\mathbb{R}^s$ is an $s$-dimensional Euclidean space.
Let $\mathbf{x} = (\mathbf{x}_{\text{time}}, \mathbf{x}_{\text{space}})^T \in \mathcal{Q}_{s,t}^\beta$, $\mathbf{x}_{\text{time}} \in \mathbb{R}^{t+1}$ and $\mathbf{x}_{\text{space}} \in \mathbb{R}^s$. The map $\psi$ and its inverse $\psi^{-1}$ are given by:
\begin{align}
    &\psi(\mathbf{x}) = \left( \sqrt{|-\beta|} \frac{\mathbf{x}_{\text{time}}}{\|\mathbf{x}_{\text{time}}\|_2}, \mathbf{x}_{\text{space}} \right) \quad (\text{assuming } \beta < 0) , \\
    &\psi^{-1}(\mathbf{U}, \mathbf{V}) = \left( \frac{\sqrt{|-\beta| + \|\mathbf{V}\|_2^2}}{\sqrt{|-\beta|}} \mathbf{U}, \mathbf{V} \right) ,
    \label{eq:psi_inv_map}
\end{align}
where $(\mathbf{U}, \mathbf{V}) \in S_t^{-\beta} \times \mathbb{R}^s$. For $\beta > 0$, similar forms exist.
\end{theorem}
This diffeomorphism allows operations to be defined in the product space $S_t^{-\beta} \times \mathbb{R}^s$, which is geodesically connected, and then mapped back to $\mathcal{Q}_{s,t}^\beta$. Operations in this product space, such as logarithmic and exponential maps, can be conveniently decomposed into operations on the spherical/hyperbolic component and the Euclidean component independently. For instance, the logarithmic map in the product space, $\log^{S_t^{-\beta} \times \mathbb{R}^s}$, applied to a point $(\mathbf{U}, \mathbf{V})$ concerning an origin $(\mathbf{o}_S, \mathbf{0}_E)$, is $(\log^{\text{sph}}_{\mathbf{o}_S}(\mathbf{U}), \mathbf{V} - \mathbf{0}_E)$. 
\begin{definition}[Diffeomorphic Logarithmic and Exponential Maps~\cite{13}]
Let $\mathbf{o}$ be a reference point (origin) on $\mathcal{Q}_{s,t}^\beta$, the diffeomorphic logarithmic map $\hat{\log}_{\mathbf{o}}: \mathcal{Q}_{s,t}^\beta \to \mathcal{T}_{\mathbf{o}}\mathcal{Q}_{s,t}^\beta$ and the diffeomorphic exponential map $\hat{\exp}_{\mathbf{o}}: \mathcal{T}_{\mathbf{o}}\mathcal{Q}_{s,t}^\beta \to \mathcal{Q}_{s,t}^\beta$ are defined as:
\begin{align}
    \hat{\log}_{\mathbf{o}}(\mathbf{y}) &= (d\psi^{-1})_{\psi(\mathbf{o})} \left( \log^{S_t^{-\beta} \times \mathbb{R}^s}_{\psi(\mathbf{o})}(\psi(\mathbf{y})) \right) ,\label{eq:diffeo_log_map} \\
    \hat{\exp}_{\mathbf{o}}(\pmb{\xi}) &= \psi^{-1} \left( \exp^{S_t^{-\beta} \times \mathbb{R}^s}_{\psi(\mathbf{o})}((d\psi)_{\mathbf{o}}(\pmb{\xi})) \right), \label{eq:diffeo_exp_map}
\end{align}
where $\mathcal{T}_{\mathbf{o}}\mathcal{Q}_{s,t}^\beta$ is the tangent space at $\mathbf{o}$, and $(d\psi)_{\mathbf{o}}$ and $(d\psi^{-1})_{\psi(\mathbf{o})}$ are the pushforward and pullback of the diffeomorphism, respectively. For a specific choice of origin $\mathbf{o}$ (e.g., the "pole" $[0, \dots, 0, \sqrt{|\beta|}]^T$ if $\beta<0$ and the last component is time-like, or similar canonical choices), these maps simplify calculations by aligning tangent spaces. In practice, computations often involve applying $\psi$, performing operations in the product space (which has simpler log/exp maps, e.g., standard spherical/Euclidean log/exp), and then applying $\psi^{-1}$. The tangent space $\mathcal{T}_{\mathbf{o}}\mathcal{Q}_{s,t}^\beta$ is a vector space, allowing for linear operations like weighted sums, which are then mapped back to the manifold.
\label{dlem}
\end{definition}
Using this diffeomorphism, the diffeomorphic logarithmic and exponential maps are defined as Definition~\ref{dlem}.
In Definition~\ref{dlem}, the tools, particularly $\psi, \psi^{-1}, \hat{\log}_{\mathbf{o}}, \hat{\exp}_{\mathbf{o}}$, and the component-wise operations on $S_t^{-\beta}$ (e.g., $\log^{\text{sph}}, \exp^{\text{sph}}$) and $\mathbb{R}^s$, form the geometric foundation of the operations in our PR-CapsNet.

\subsection{Capsule Networks and Dynamic Routing}
\label{subsec:capsule_networks}
Capsule Networks (CapsNets)~\cite{45} were introduced as an alternative to traditional Convolutional Neural Networks (CNNs) to better model hierarchical relationships and improve robustness to viewpoint changes. Instead of scalar activations, CapsNets use vectors (or matrices) called "capsules" to represent entities or their parts. The length of a capsule's activity vector represents the probability of an entity's presence, while its orientation encodes the entity's instantiation parameters (e.g., pose, deformation).

A core component of CapsNets is the \emph{dynamic routing} mechanism, which allows capsules in a lower layer to iteratively decide how to send their output to capsules in a higher layer. This process enables the network to group capsules that form coherent parts of a larger entity.
The standard dynamic routing algorithm between a layer of children capsules $\{\mathbf{u}_i\}$ and a layer of parent capsules $\{\mathbf{v}_j\}$ typically involves the following steps with a fixed number of iterations. \textbf{Step 1 (Prediction Vector Generation):} Each child capsule $i$ generates a prediction vector $\hat{\mathbf{u}}_{j|i}$ for each potential parent capsule $j$. This is usually done by multiplying the child capsule's output $\mathbf{u}_i$ by a trainable weight matrix $\mathbf{W}_{ij}$:
\begin{equation}
    \hat{\mathbf{u}}_{j|i} = \mathbf{W}_{ij} \mathbf{u}_i.
    \label{eq:caps_prediction_euclidean}
\end{equation}
These prediction vectors represent the "vote" of child $i$ for the instantiation parameters of parent $j$. 
\textbf{Step 2 (Coupling Coefficient Update):}  The strength of connection, or routing, from child $i$ to parent $j$ is determined by coupling coefficients $c_{ij}$. These are computed by applying a softmax function to log prior probabilities $b_{ij}$:
\begin{equation}
    c_{ij} = \frac{\exp(b_{ij})}{\sum_k \exp(b_{ik})}.
    \label{eq:caps_coupling_coeffs}
\end{equation}
Initially, all $b_{ij}$ are set to zero. 
\textbf{Step 3 (Weighted Aggregation) :} The input to a parent capsule $j$, denoted $\mathbf{s}_j$, is a weighted sum of all prediction vectors $\hat{\mathbf{u}}_{j|i}$ from the children capsules connected to it:
\begin{equation}
    \mathbf{s}_j = \sum_i c_{ij} \hat{\mathbf{u}}_{j|i}.
    \label{eq:caps_weighted_sum_euclidean}
\end{equation}
\textbf{Step 4 (Squashing):} The output $\mathbf{v}_j$ of parent capsule $j$ is obtained by applying a non-linear "squashing" function to $\mathbf{s}_j$. This function scales short vectors to almost zero length and long vectors to a length slightly below 1, preserving their orientation:
\begin{equation}
    \mathbf{v}_j = \frac{\|\mathbf{s}_j\|^2}{1 + \|\mathbf{s}_j\|^2} \frac{\mathbf{s}_j}{\|\mathbf{s}_j\|}.
    \label{eq:caps_squashing_euclidean}
\end{equation}
\textbf{Step 5 (Agreement and Logit Update):} The log priors $b_{ij}$ are then updated based on the agreement between the current output of parent capsule $j$ ($\mathbf{v}_j$) and the prediction made by child capsule $i$ ($\hat{\mathbf{u}}_{j|i}$). Agreement is typically measured by a dot product:
\begin{equation}
    b_{ij} \leftarrow b_{ij} + \mathbf{v}_j \cdot \hat{\mathbf{u}}_{j|i}.
    \label{eq:caps_agreement_update_euclidean}
\end{equation}
    
\textbf{Steps 2-5} are repeated for a fixed number of routing iterations. This iterative process allows the network to dynamically route information and form strong connections between capsules that agree.
While effective, these operations are defined in Euclidean space. Adapting CapsNets to non-Euclidean manifolds, such as the pseudo-hyperboloid, requires re-formulating these steps to be compatible with the underlying geometry. For example, HYPERCAPS~\cite{42} adapted these principles to hyperbolic space using M\"obius gyrovector operations, where Möbius addition replaced vector summation (Eq. (\ref{eq:caps_weighted_sum_euclidean})) and Möbius matrix-vector multiplication generated predictions (Eq. (\ref{eq:caps_prediction_euclidean})), naturally keeping capsules within the hyperbolic ball without explicit squashing.

\section{PR-CapsNet}
\label{sec:method}
In this section, we introduce our PR-CapsNet in detail with the help of the foundational concepts from Pseudo-Riemannian geometry and capsule networks in Section~\ref{sec:preliminary}. Compared to the original CapsNet, the core innovations of PR-CapsNet are introducing an Adaptive Curvature Routing (ACR) module based on Pseudo-Riemannian Tangent Space Routing and a Pseudo-Riemannian Capsule Classifier (PRCC). The ACR allows PR-CapsNet to dynamically learn and leverage features processed through different geometric perspectives, enhancing its ability to model graphs
with diverse and varying local geometric characteristics. The PRCC projects capsule embeddings to tangent spaces and uses curvature-weighted softmax for classification while preserving geometric properties.

\subsection{Pseudo-Riemannian Capsule Routing}
\label{sec:pcr}

\noindent\textbf{Geometric Semantics.} 
In our framework, the decomposition into spherical and Euclidean subspaces is not arbitrary. The \textit{time-like} dimensions (associated with the negative signature in the metric) naturally model hyperbolic geometry properties, making them suitable for capturing hierarchical organizations and causal dependencies in the graph. Conversely, the \textit{space-like} dimensions (positive signature) behave like Euclidean or Spherical geometry, excelling at modeling cyclic structures and diverse community clusters. By utilizing diffeomorphic transformations, PR-CapsNet effectively leverages this duality to represent mixed graph topologies within a unified manifold.
Conventional dynamic routing relies on linear operations incompatible with pseudo-Riemannian geometry. Geodesic disconnectedness on $\mathcal{Q}_{\beta}^{s,t}$ invalidates direct aggregation when $\langle x,y\rangle_{\text{ps}} \geq |\beta|$, causing $\log_x(y)$ to become undefined. Following Pseudo-Riemannian GCN~\cite{13}, we address non-Euclidean nature and geodesic disconnectedness by using a diffeomorphism $\psi: \mathcal{Q}_\beta^{s,t} \to S_t^{-\beta} \times \mathbb{R}^s$, which can preserve metric structure while decomposing the pseudo-hyperboloid into spherical and Euclidean components. This enables geometric operations in tangent space $\mathcal{T}_{\mathbf{o}}\mathcal{Q}_{s,t}^\beta$ using diffeomorphic logarithmic map $\hat{\log}_{\mathbf{o}}$ and exponential map $\hat{\exp}_{\mathbf{o}}$ that maintains geometric consistency.
Given each child capsule $i$ (state $\mathbf{u}_i \in \mathcal{Q}_{s_{\text{in}}, t_{\text{in}}}^{\beta_{\text{in}}}$) and parent capsule $j$ (state $\mathbf{v}_j \in \mathcal{Q}_{s_{\text{out}},t_{\text{out}}}^{\beta_{\text{out}}}$), the dynamic routing with pseudo-Riemannian geometry involves the following steps. 

The \textbf{Step 1 (Prediction Vector Generation)} is that child capsule $i$ generates prediction $\hat{\mathbf{u}}_{j|i}$ for parent $j$ via geometric transformation in decomposed subspaces:
    \begin{equation}
        \hat{\mathbf{u}}_{j|i} = \psi^{-1} \left( \exp^{\text{sph}}_{S_{t_{\text{out}}}}\left(\mathbf{W}^{\text{sph}}_{ij} \log^{\text{sph}}_{S_{t_{\text{in}}}}(\mathbf{u}_{i}^{\text{sph}})\right) \parallel \mathbf{W}^{\text{euc}}_{ij} \mathbf{u}_{i}^{\text{euc}} \right),
        \label{eq:prediction_vector}
    \end{equation}
    where $\mathbf{u}_i = (\mathbf{u}_{i}^{\text{sph}}, \mathbf{u}_{i}^{\text{euc}})$ via $\psi$, with $\mathbf{u}_{i}^{\text{sph}} \in S_{t_{\text{in}}}^{-\beta_{\text{in}}}$ and $\mathbf{u}_{i}^{\text{euc}} \in \mathbb{R}^{s_{\text{in}}}$. Trainable matrices $\mathbf{W}^{\text{sph}}_{ij} \in \mathbb{R}^{t_{\text{out}} \times t_{\text{in}}}$ and $\mathbf{W}^{\text{euc}}_{ij} \in \mathbb{R}^{s_{\text{out}} \times s_{\text{in}}}$ transform spherical and Euclidean components. This diffeomorphic transformation preserves pseudo-Riemannian distance structure as $\psi$ acts as an isometry.
The \textbf{Step 2 (Weighted Aggregation in Tangent Space)} is that prediction vectors are aggregated geometrically by: (i) mapping to common tangent space $\mathcal{T}_{\mathbf{o}}\mathcal{Q}_{s_{\text{out}},t_{\text{out}}}^{\beta_{\text{out}}}$ via $\hat{\log}_{\mathbf{o}}$; (ii) weighted summation with routing coefficients $c_{ij}$; (iii) mapping back via $\hat{\exp}_{\mathbf{o}}$:
    \begin{equation}
        \mathbf{s}_j = \hat{\exp}_{\mathbf{o}} \left( \sum_i c_{ij} \hat{\log}_{\mathbf{o}}(\hat{\mathbf{u}}_{j|i}) \right),
        \label{eq:weighted_aggregation}
    \end{equation}
approximating the Fréchet mean on the manifold.
The \textbf{Step 3 (Non-linear Activation and Projection)} is that the aggregated vector $\mathbf{s}_j$ undergoes: (i) tangent space mapping via $\hat{\log}_{\mathbf{o}}$; (ii) element-wise activation $\sigma$; (iii) manifold reconstruction with projection:
    \begin{equation}
        \mathbf{v}_j = \text{Proj}_{\mathcal{Q}_{s_{\text{out}},t_{\text{out}}}^{\beta_{\text{out}}}} \left( \hat{\exp}_{\mathbf{o}} \left( \sigma \left( \hat{\log}_{\mathbf{o}}(\mathbf{s}_j) \right) \right) \right),
        \label{eq:activation_projection}
    \end{equation}
    ensuring $\mathbf{v}_j$ remains on the manifold with numerical stability.
The \textbf{Step 4 (Dynamic Routing Update)} is that coupling coefficients $c_{ij}$ are refined through tangent space agreement:
    \begin{align}
        b_{ij} &\leftarrow b_{ij} + \langle \hat{\log}_{\mathbf{o}}(\mathbf{v}_j), \hat{\log}_{\mathbf{o}}(\hat{\mathbf{u}}_{j|i}) \rangle_{\text{ps}}, \\
        c_{ij} &= \text{softmax}_j(b_{ij}).
    \end{align}
    This ensures that capsules with aligned predictions receive higher coupling weights.

\subsection{Adaptive Curvature Routing}
\label{sec:acr}
While the dynamic routing with pseudo-Riemannian geometry enables operations on a manifold with a fixed curvature $\beta$, real-world data often exhibits complex structures that should be best modeled by spaces with varying local geometric properties. Although the pseudo-hyperboloid $\mathcal{Q}_{s,t}^\beta$ with a fixed scalar $\beta$ provides a specific inductive bias (hyperbolic-like for $\beta<0$, spherical-like for $\beta>0$). This is insufficient for graphs simultaneously exhibiting strong hierarchical structures (negative curvature) and dense community clusters (positive curvature). To model such complex topologies, the network needs to leverage different geometric properties for different aspects or dimensions of the features during routing. Therefore, the ACR module extends the routing mechanism to handle these geometric characteristics adaptively within the same pseudo-Riemannian space.

\subsubsection{Geometric Perspectives and Learned Weights.}
Instead of attempting to define a dynamically changing metric tensor within a layer, we frame the adaptive curvature concept as learning to fuse information processed through different \emph{geometric perspectives} or \emph{feature interpretations}. We introduce $K$ such perspectives. Each perspective $k \in \{1, \dots, K\}$ is associated with a specific interpretation of the children capsule features, potentially emphasizing aspects best described by different curvatures $\beta_k$. These perspectives are captured by learned linear transformations $\mathbf{W}^{\text{sph}}_{ij,k}$ and $\mathbf{W}^{\text{euc}}_{ij,k}$ that produce \emph{perspective-specific prediction vectors}.
For each child capsule $i$, each parent capsule $j$, and each perspective $k$, a perspective-specific prediction vector $\hat{\mathbf{u}}_{j|i,k}$ is generated according to Eq. (\ref{eq:prediction_vector}) with transformations specific to perspective $k$:
\begin{equation}
    \hat{\mathbf{u}}_{j|i,k} = \psi^{-1} \left( \exp^{\text{sph}}_{S_{t_{\text{out},k}}}\left(\mathbf{W}^{\text{sph}}_{ij,k} \log^{\text{sph}}_{S_{t_{\text{in}}}}(\mathbf{u}_{i}^{\text{sph}})\right) \parallel \mathbf{W}^{\text{euc}}_{ij,k} \mathbf{u}_{i}^{\text{euc}} \right).
    \label{eq:perspective_prediction_vector}
\end{equation}
where $\mathbf{u}_i = (\mathbf{u}_{i}^{\text{sph}}, \mathbf{u}_{i}^{\text{euc}})$ is the decomposition of $\mathbf{u}_i$ via $\psi$, with $\mathbf{u}_{i}^{\text{sph}} \in S_{t_{\text{in}}}^{-\beta_{\text{in}}}$ and $\mathbf{u}_{i}^{\text{euc}} \in \mathbb{R}^{s_{\text{in}}}$. $\mathbf{W}^{\text{sph}}_{ij,k} \in \mathbb{R}^{t_{\text{out},k} \times t_{\text{in}}}$ and $\mathbf{W}^{\text{euc}}_{ij,k} \in \mathbb{R}^{s_{\text{out},k} \times s_{\text{in}}}$ are trainable transformation matrices for perspective $k$. The dimensions $t_{\text{out},k}, s_{\text{out},k}$ and the target sphere curvature might vary per perspective or layer, allowing flexibility.

\subsubsection{Curvature-Aware Gating Mechanism.}
The key insight for adaptive curvature routing is that different geometric perspectives should be selected based on the \emph{intrinsic geometric properties} of the capsule relationships rather than arbitrary feature comparisons. Each perspective $k$ with associated curvature parameter $\beta_k$ should be weighted according to how well its geometric assumptions align with the local manifold structure of the data relationship between capsules $i$ and $j$.
This geometric suitability is determined through three complementary measures: 1) \textbf{Curvature Compatibility}, which measures the match degree between the perspective's curvature $\beta_k$ and the estimated local curvature of the capsule relationship; 2) \textbf{Feature Alignment}, which measures the alignment degree among the capsule features with the perspective's geometric structure in the tangent space; 3) \textbf{Routing Consistency}, which measures the consistent degree between this perspective selection and the overall routing dynamics.
Therefore, we introduce curvature-aware gating weights $\gamma_{ij,k} \in [0,1]$ that modulate the importance of the prediction from children capsule $i$ to parent capsule $j$ when viewed through geometric perspective $k$. These weights are computed based on the geometric compatibility between the capsule relationship and the perspective's assumptions:

\begin{equation}
    \gamma_{ij,k} = \sigma \left( \alpha_{\text{curv}} \cdot \mathcal{C}_{ij,k} + \alpha_{\text{align}} \cdot \mathcal{A}_{ij,k} + \alpha_{\text{route}} \cdot \mathcal{R}_{ij,k} \right),
    \label{eq:gating_weights_new}
\end{equation}
where $\alpha_{\text{curv}}, \alpha_{\text{align}}, \alpha_{\text{route}} \in \mathbb{R}$ are learnable importance weights, $\sigma$ is the sigmoid function, and the three terms are defined as follows:

\textbf{Curvature Compatibility Term $\mathcal{C}_{ij,k}$} measures how well perspective $k$'s curvature parameter matches the estimated local curvature:
\begin{equation}
    \mathcal{C}_{ij,k} = -\frac{1}{2}\left|\hat{\kappa}_{ij} - \beta_k\right|^2,
    \label{eq:curvature_compatibility}
\end{equation}
where $\hat{\kappa}_{ij}$ is the estimated local curvature of the relationship between capsules $i$ and $j$, computed in the tangent space as:
\begin{equation}
    \hat{\kappa}_{ij} = \frac{\langle \hat{\log}_{\mathbf{o}}(\mathbf{u}_i), \hat{\log}_{\mathbf{o}}(\mathbf{v}_j^{\text{prev}}) \rangle}{\|\hat{\log}_{\mathbf{o}}(\mathbf{u}_i)\|_2 \cdot \|\hat{\log}_{\mathbf{o}}(\mathbf{v}_j^{\text{prev}})\|_2},
    \label{eq:local_curvature}
\end{equation}
where $\mathbf{v}_{j}^{\text{prev}}$ is the current state of parent capsule $j$ from the previous routing iteration, $\langle \cdot, \cdot \rangle$ denotes the inner product in the tangent space, and $\|\cdot\|_2$ is the Euclidean norm.

\textbf{Feature Alignment Term $\mathcal{A}_{ij,k}$} captures how well the capsule features align with perspective $k$'s geometric structure:
\begin{equation}
    \mathcal{A}_{ij,k} = \mathbf{w}_k^T \tanh\left(\mathbf{W}^{\text{align}}_k \left[ \hat{\log}_{\mathbf{o}}(\mathbf{u}_i) \parallel \hat{\log}_{\mathbf{o}}(\mathbf{v}_j^{\text{prev}}) \right]\right),
    \label{eq:feature_alignment}
\end{equation}
where $\mathbf{w}_k \in \mathbb{R}^{d_{\text{align}}}$ and $\mathbf{W}^{\text{align}}_k \in \mathbb{R}^{d_{\text{align}} \times 2d_{\text{tangent}}}$ are learnable parameters specific to perspective $k$, and $d_{\text{tangent}}$ is the dimension of the tangent space.

\textbf{Routing Consistency Term $\mathcal{R}_{ij,k}$} ensures consistency with the overall routing pattern:
\begin{equation}
    \mathcal{R}_{ij,k} = \log c_{ij} \cdot \mathbf{e}_k^T \mathbf{W}^{\text{C}} \mathbf{h}_{ij},
    \label{eq:routing_consistency}
\end{equation}
where $c_{ij}$ is the coupling coefficient from the agreement mechanism, $\mathbf{e}_k \in \{0,1\}^K$ is a one-hot encoding for perspective $k$, $\mathbf{W}^{\text{C}} \in \mathbb{R}^{K \times d_{\text{CTX}}}$ is a learnable matrix, and $\mathbf{h}_{ij} \in \mathbb{R}^{d_{\text{CTX}}}$ captures the routing context between capsules $i$ and $j$.

Eq. (\ref{eq:gating_weights_new}) provides clear geometric intuition for perspective selection. First, the curvature compatibility term $\mathcal{C}_{ij,k}$ directly measures how well perspective $k$'s geometric assumptions (encoded in $\beta_k$) match the estimated local geometry of the capsule relationship. Perspectives with curvature parameters closer to the estimated local curvature $\hat{\kappa}_{ij}$ receive exponentially higher weights.
Second, the local curvature estimation in Eq. (\ref{eq:local_curvature}) uses the normalized dot product in tangent space, which has a clear geometric interpretation to measure the alignment between capsule representations in the manifold's local coordinate system.
Third, the multi-component formulation allows the model to balance geometric compatibility with feature-specific alignment and routing dynamics, providing a robust and interpretable selection mechanism.

\subsubsection{Dynamic Aggregation of Multi-Perspective Features.}
The final aggregated vector $\mathbf{s}_j$ for parent capsule $j$ is computed by aggregating the perspective-specific prediction vectors $\hat{\mathbf{u}}_{j|i,k}$ from all child capsules $i$. This geometric aggregation operates in the tangent space using $\hat{\log}_{\mathbf{o}}$ and $\hat{\exp}_{\mathbf{o}}$ operations. Specifically, each prediction $\hat{\mathbf{u}}_{j|i,k}$ first undergoes logarithmic mapping to the tangent space $\hat{\log}_{\mathbf{o}}(\hat{\mathbf{u}}_{j|i,k})$. Then, composite weighting is applied to standard coupling coefficients $c_{ij}$ from the agreement mechanism and learned perspective gates $\gamma_{ij,k}$ for connection $i \rightarrow j$ under perspective $k$. Subsequently, weighted tangent vectors are linearly combined as follows:
\begin{equation}
    \sum_{i=1}^N \sum_{k=1}^K c_{ij} \gamma_{ij,k} \hat{\log}_{\mathbf{o}}(\hat{\mathbf{u}}_{j|i,k})
\end{equation}
Finally, the final exponential mapping returns to the manifold space as follows:
\begin{equation}
    \mathbf{s}_j = \hat{\exp}_{\mathbf{o}} \left( \sum_{i=1}^N \sum_{k=1}^K c_{ij} \gamma_{ij,k} \hat{\log}_{\mathbf{o}}(\hat{\mathbf{u}}_{j|i,k}) \right)
    \label{eq:adaptive_aggregation}
\end{equation}

Eq. (\ref{eq:adaptive_aggregation}) preserves geometric consistency through two key properties: 1) \textbf{Diffeomorphic alignment} that all $\hat{\mathbf{u}}_{j|i,k}$ follow the same generative process as the base prediction vectors and 2) \textbf{Space compatibility} that the composite weights $c_{ij}\gamma_{ij,k}$ maintain both routing probabilities ($\sum_{i,k} c_{ij}\gamma_{ij,k} = 1$) and tangent space linearity. Thus,  the resulting representation $\mathbf{s}_j$ remains geometrically valid on the target manifold while achieving enhanced expressiveness through multi-perspective integration. The adaptive curvature mechanism enables the model to automatically select the most appropriate geometric interpretation for each capsule relationship. This leads to more flexible and expressive representations that handle complex data structures exhibiting mixed geometric properties.

\subsection{Pseudo-Riemannian Capsule Classifier}
\label{sec:prcc}

After obtaining the final layer of parent capsule representations through the adaptive curvature routing mechanisms, the final step is to classify these representations. 
Standard classifiers like Softmax operate on vectors in an Euclidean vector space. Applying them directly to the parent capsule states $\mathbf{v}_j$ residing on a pseudo-Riemannian manifold presents metric incompatibilities. Distances and dot products in the tangent space are the natural measures of similarity and alignment on the manifold, but they are not standard Euclidean operations on the raw capsule vectors. Therefore, the parent capsule states must be mapped appropriately or aligned to a suitable space for classification.

We address this challenge by performing the final classification in a tangent space, leveraging the tangent space representation's vector space properties while preserving metric relationships via the logarithmic map. This strategy involves mapping the final layer parent capsule states $\mathbf{v}_j^{(L)}$ to the tangent space $\mathcal{T}_{\mathbf{o}}\mathcal{Q}_{s_L,t_L}^{\beta_L}$ at the reference point $\mathbf{o}$ using the diffeomorphic logarithmic map $\hat{\log}_{\mathbf{o}}$. This provides a vector representation $\mathbf{z}_j$ suitable for linear operations, for each parent capsule $j$ in the final layer:
\begin{equation}
    \mathbf{z}_j = \hat{\log}_{\mathbf{o}} \left( \mathbf{v}_j^{(L)} \right) \in \mathcal{T}_{\mathbf{o}}\mathcal{Q}_{s_L,t_L}^{\beta_L}.
    \label{eq:tangent_space_mapping}
\end{equation}
where $\mathbf{v}_j^{(L)}$ is the state of the final layer parent capsule $j$. 
The similarity or "alignment" between two parent capsule states $\mathbf{x}$ and $\mathbf{y}$ on the manifold can then be naturally measured by the pseudo-Riemannian inner product of their tangent vectors in $\mathcal{T}_{\mathbf{o}}\mathcal{Q}_{s,t}^\beta$:
\begin{equation}
    \text{Align}(\mathbf{x},\mathbf{y}) = \frac{\langle \hat{\log}_{\mathbf{o}}(\mathbf{x}), \hat{\log}_{\mathbf{o}}(\mathbf{y}) \rangle_{\text{ps}}}{\|\hat{\log}_{\mathbf{o}}(\mathbf{x})\|_{\text{ps}} \|\hat{\log}_{\mathbf{o}}(\mathbf{y})\|_{\text{ps}}}.
    \label{eq:alignment}
\end{equation}
where $\|\mathbf{v}\|_{\text{ps}} = \sqrt{|\langle \mathbf{v}, \mathbf{v} \rangle_{\text{ps}}|}$ is the pseudo-Riemannian norm of a tangent vector $\mathbf{v}$. This preserves pseudo-Riemannian inner product properties through the logarithmic mapping, providing a geometrically meaningful similarity score.

\begin{algorithm}[t] 
\caption{The Pseudo-Riemannian Routing Algorithm}
\label{alg:PR-CapsNet_routing}
\begin{algorithmic}[1]
\Procedure{PRR\_ROUTING}{$T$, $\{\mathbf{u}_i\}_{i=1}^{N_c}$,} 
\State \hspace{\algorithmicindent} $\{\mathbf{W}^{\text{sph}}_{ij,k}, \mathbf{W}^{\text{euc}}_{ij,k}, \mathbf{W}^{\text{gate}}_{ij,k}\}$, $\mathbf{o}$,
\State \hspace{\algorithmicindent} $\psi, \psi^{-1}, \hat{\log}_{\mathbf{o}}, \hat{\exp}_{\mathbf{o}}, \exp^{\text{sph}}, \log^{\text{sph}}, \sigma, \langle \cdot, \cdot \rangle_{\text{ps}}$ 
% ...
\State Initialize $b_{ij} \leftarrow 0$ for all children $i$ and parent $j$
\State Let $\mathbf{v}_j^{\text{prev}} \leftarrow \mathbf{o}$ for all parent capsules $j$ 
\For {$t = 1$ to $T$}
    \State $c_{ij} \leftarrow \frac{\exp(b_{ij})}{\sum_{j'} \exp(b_{ij'})}$ 
    \State $\mathbf{S}^{\tau}_j \leftarrow \mathbf{0}$ for all parent $j$
    \For {each children capsule $i$}
        \State $(\mathbf{u}_i^{\text{sph}}, \mathbf{u}_i^{\text{euc}}) \leftarrow \psi(\mathbf{u}_i)$ 
        \For {each parent capsule $j$}
            \For {each perspective $k = 1$ to $K$}
                \State $\hat{\mathbf{u}}_{j|i,k}^{\text{sph}} \leftarrow \exp^{\text{sph}}\left(\mathbf{W}^{\text{sph}}_{ij,k} \log^{\text{sph}}(\mathbf{u}_i^{\text{sph}})\right)$ 
                \State $\hat{\mathbf{u}}_{j|i,k} \leftarrow \psi^{-1} \left( \hat{\mathbf{u}}_{j|i,k}^{\text{sph}} \parallel \mathbf{W}^{\text{euc}}_{ij,k} \mathbf{u}_i^{\text{euc}} \right)$ 
                \State $\gamma_{ij,k} \leftarrow \sigma \left( \mathbf{W}^{\text{gate}}_{ij,k} \left[ \hat{\log}_{\mathbf{o}}(\mathbf{u}_i) \right. \right.$ 
                \State \hspace{\algorithmicindent} $\left. \left. \parallel \hat{\log}_{\mathbf{o}}(\mathbf{v}_j^{\text{prev}}) \right] \right)$ 
                \State $\hat{\pmb{\xi}}_{j|i,k} \leftarrow \hat{\log}_{\mathbf{o}}(\hat{\mathbf{u}}_{j|i,k})$ 
                \State $\mathbf{S}^{\tau}_j \leftarrow \mathbf{S}^{\tau}_j + c_{ij} \gamma_{ij,k} \hat{\pmb{\xi}}_{j|i,k}$ 
            \EndFor
        \EndFor
    \EndFor
    \State $\mathbf{s}_j \leftarrow \hat{\exp}_{\mathbf{o}} \left( \mathbf{S}^{\tau}_j \right)$
    \State $\mathbf{v}_j \leftarrow \hat{\exp}_{\mathbf{o}}(\sigma(\hat{\log}_{\mathbf{o}}(\mathbf{s}_j)))$ 
    \State $\mathbf{v}_j^{\text{prev}} \leftarrow \mathbf{v}_j$
    \For {each children $i$ and parent $j$}
        \For {each perspective $k = 1$ to $K$}
            \State $b_{ij} \leftarrow b_{ij} + \gamma_{ij,k} \langle \hat{\log}_{\mathbf{o}}(\mathbf{v}_j), \hat{\log}_{\mathbf{o}}(\hat{\mathbf{u}}_{j|i,k}) \rangle_{\text{ps}}$ 
        \EndFor
    \EndFor
\EndFor
\State \textbf{return} $\{\mathbf{v}_j\}_{j=1}^{N_p}$ 
\EndProcedure
\end{algorithmic}
\end{algorithm}

\subsection{Pseudo-Riemannian Routing Algorithm}
\label{sec:activation}
Non-linear activation functions in PR-CapsNet are applied to vectors in the tangent space $\mathcal{T}_{\mathbf{o}}\mathcal{Q}_{s,t}^\beta$. Standard activation functions like ReLU or tanh can be applied element-wise to the tangent vector before the final exponential map back to the manifold. For instance, the activation function $\sigma(\cdot)$ in Eq. (\ref{eq:activation_projection}) is applied to the tangent vector $\hat{\log}_{\mathbf{o}}(\mathbf{s}_j)$. This is a common practice in geometric deep learning on curved spaces, providing non-linearity while operating in a vector space.
Overall, the Pseudo-Riemannian Routing (\textbf{Algorithm~\ref{alg:PR-CapsNet_routing}}) of the PR-CapsNet procedure is detailed below. This algorithm integrates the PCR and ACR mechanisms described in Section \ref{sec:method}. The routing process iteratively refines coupling coefficients $c_{ij}$. Parent capsule states $\mathbf{v}_j$ are updated based on agreement between predicted and aggregated states, using geometric operations adapted for the pseudo-Riemannian manifold $\mathcal{Q}_{s,t}^\beta$ and its tangent space $\mathcal{T}_\mathbf{o}\mathcal{Q}_{s,t}^\beta$.
The algorithm takes as input the children capsule states $\{\mathbf{u}_i\}_{i=1}^{N_c}$ where $\mathbf{u}_i \in \mathcal{Q}_{s_{\text{in}}, t_{\text{in}}}^{\beta_{\text{in}}}$, along with transformation matrices $\{\mathbf{W}^{\text{sph}}_{ij,k}, \mathbf{W}^{\text{euc}}_{ij,k}, \mathbf{W}^{\text{gate}}_{ij,k}\}$ for all children $i$, parent $j$, and perspective $k$, and the number of routing iterations $T$. It outputs the parent capsule states $\{\mathbf{v}_j\}_{j=1}^{N_p}$ where $\mathbf{v}_j \in \mathcal{Q}_{s_{\text{out}}, t_{\text{out}}}^{\beta_{\text{out}}}$. Throughout the algorithm, we employ several key geometric operations: the diffeomorphism mapping $\psi, \psi^{-1}$ that establishes a correspondence between $\mathcal{Q}_{s,t}^{\beta}$ and $S_{t}^{-\beta} \times \mathbb{R}^s$; the diffeomorphic logarithmic and exponential maps $\hat{\log}_{\mathbf{o}}, \hat{\exp}_{\mathbf{o}}$ that facilitate movement between the manifold and its tangent space; the standard exponential and logarithmic maps $\exp^{\text{sph}}, \log^{\text{sph}}$ on spheres (or hyperbolic spaces); and the pseudo-Riemannian inner product $\langle\cdot,\cdot\rangle_{\text{ps}}$ for computing agreement.

The computational complexity of PR-CapsNet is primarily dominated by the Pseudo-Riemannian Routing mechanism. For a layer with $N$ input capsules, $M$ parent capsules, $K$ geometric perspectives, and $T$ routing iterations, the complexity is approximately $O(T \cdot K \cdot |E|_{caps} \cdot d^2)$, where $|E|_{caps}$ represents the number of connections between child and parent capsules and $d$ is the capsule dimension. Although the diffeomorphic mappings ($\log_o, \exp_o$) introduce a constant computational overhead, the operations are parallelizable across perspectives. Compared to standard MPNNs ($O(|E|_{graph} \cdot d^2)$), the cost increases linearly with $K$ and $T$. As evidenced by the runtime analysis in Table~\ref{tab:graph_classification_results_std_3dp}, PR-CapsNet remains scalable for mid-sized graphs comparable to other advanced geometric GNNs.

\begin{table}[h]
\setlength{\tabcolsep}{1mm}
\centering
\resizebox{1\linewidth}{!}{
\begin{tabular}{l|l|ccccc}
\hline
 \multirow{5}{*}{\makecell{ Node \\ Classification }  } & Dataset  & Nodes & Edges & Features & Classes \\ \cline{2-6}
 &Cora\cite{46}  & 2,708 & 5,278 & 1,433 & 7 \\
&Citeseer\cite{47}  & 3,327 & 4,676 & 3,703 & 6 \\
&PubMed\cite{48}  & 19,717 & 44,338 & 500 & 3 \\
&CoauthorCS\cite{49}  & 18,333 & 81,894 & 6,805 & 15 \\
\hline \hline
 \multirow{6}{*}{\makecell{ Graph \\ Classification }  }  & Dataset  & Graphs & Nodes (avg) & Edges (avg) & Classes \\ \cline{2-6}
& MUTAG\cite{50}  & 188 & 17.9 & 19.8  & 2 \\
& NCI1\cite{51}  & 4,110 & 29.8 & 32.3  & 2 \\
& NCI109\cite{51}  & 4,127 & 29.6 & 32.1  & 2 \\
& DD\cite{52}  & 1,178 & 284.3 & 715.7  & 2 \\
& PROTEINS\cite{53}  & 1,113 & 39.1 & 72.8 & 2 \\
\hline
\end{tabular}}
\caption{Datasets for Node and Graph Classification.}
\label{tab:node_datasets}
\end{table}

\begin{table*}[t]
  \centering
  \resizebox{\linewidth}{!}{%
    \begin{tabular}{|l|ccc|ccc|ccc|ccc|ccc| }
      \hline
      \multirow{2}{*}{Method} & 
      \multicolumn{3}{c|}{DD} &
      \multicolumn{3}{c|}{MUTAG} &
      \multicolumn{3}{c|}{NCI1} &
      \multicolumn{3}{c|}{NCI109} &
      \multicolumn{3}{c|}{PROTEINS} \\
      \cline{2-16}
      & Param & Runtime & ACC & Param & Runtime & ACC & Param & Runtime & ACC & Param & Runtime & ACC & Param & Runtime & ACC\\ 
      \hline
      DGCNN~\cite{54} & 4,874 & 194.84 & 0.722 $\pm$ 0.020 & 3,562 & 8.06 & 0.715 $\pm$ 0.041 & 4,042 & 166.30 & 0.673 $\pm$ 0.014 & 4,058 & 166.46 & 0.693 $\pm$ 0.004 & 3498 & 60.44 & 0.727 $\pm$ 0.013 \\
      GAT~\cite{55} & 6,994 & 18.04 & 0.614 $\pm$ 0.036 & 1,746 & 2.10& 0.613 $\pm$ 0.018 & 3,666 & 31.95 & 0.530 $\pm$ 0.014 & 3,730 & 32.14 & 0.519 $\pm$ 0.023 & 1,554 & 9.76 & 0.478 $\pm$ 0.029 \\
      GATv2~\cite{56} & 6,994 & 18.17 & 0.609 $\pm$ 0.049 & 1,746 & 2.27 & 0.613 $\pm$ 0.037 & 3,666 & 33.07 & 0.530 $\pm$ 0.014 & 3,730 & 32.82 & 0.519 $\pm$ 0.032 & 1,554 & 10.32 & 0.480 $\pm$ 0.062 \\
      GraphGPS~\cite{57} & 133,050 & 86.51 & 0.696 $\pm$ 0.093 & 129,434 & 7.24 & 0.807 $\pm$ 0.032 & 130,554 & 47.19 & 0.675 $\pm$ 0.009 & 130,602 & 47.30 & 0.599 $\pm$ 0.067 & 128,922 & 42.42 & 0.668 $\pm$ 0.023 \\
      SSGClassifier~\cite{59} & 14,210 & 17.46 & 0.264 $\pm$ 0.112 & 8,962 & 2.01 & 0.593 $\pm$ 0.001 & 10,882 & 21.17 & 0.552 $\pm$ 0.164 & 10,946 & 21.34 & 0.396 $\pm$ 0.107 & 8,706 & 9.99 & 0.196 $\pm$ 0.041 \\
      GRDL~\cite{wang2024graph} & 28,129 & 210.85 & 0.689 $\pm$ 0.049 & 11,169 & 95.41 & 0.829 $\pm$ 0.031 & 12,769 & 59.79 & 0.701 $\pm$ 0.018 & 12,737 & 60.08 & 0.725 $\pm$ 0.006 & 12,449 & 17.69 & 0.766 $\pm$ 0.023 \\
      GRATIN~\cite{abbahaddougraph} & 5,124 & 218.05 & 0.637 $\pm$ 0.041 & 2,500 & 2.53 & 0.738 $\pm$ 0.028 & 3,460 & 46.64 & 0.636 $\pm$ 0.021 & 3,492 & 49.90 & 0.622 $\pm$ 0.023 & 2,372 & 13.54 & 0.658 $\pm$ 0.025 \\
       PR-CapsNet & 7,534 & 14.35 & \textbf{0.733 $\pm$ 0.089} & 2,286 & 5.04 & \textbf{0.832 $\pm$ 0.023} & 4,206 & 54.89 & \textbf{0.711 $\pm$ 0.003} & 4270 & 55.71 & \textbf{0.732 $\pm$ 0.033} & 2,030 & 15.46 & \textbf{0.768 $\pm$ 0.010} \\
      \hline
    \end{tabular}%
  }  
  \caption{Performances on graph classification benchmarks. The unit of Runtime is seconds.}
  \label{tab:graph_classification_results_std_3dp}
\end{table*}

\begin{table*}[t]
  \centering
  \resizebox{1.0\linewidth}{!}{%
    \begin{tabular}{|l|ccc|ccc|ccc|ccc|}
      \hline
      \multirow{2}{*}{Method} & 
      \multicolumn{3}{c|}{Cora} &
      \multicolumn{3}{c|}{Citeseer} &
      \multicolumn{3}{c|}{PubMed} &
      \multicolumn{3}{c|}{CoauthorCS} \\
      \cline{2-13}
      & Param & Runtime & ACC & Param & Runtime & ACC & Param & Runtime & ACC & Param & Runtime & ACC \\ 
      \hline
      DGCNN~\cite{54} & 26,423 & 25.18 & 0.373 $\pm$ 0.045 & 62,734 & 33.23 & 0.274 $\pm$ 0.029 & 11,459 & 118.00 & 0.528 $\pm$ 0.089 & 112,447 & 205.70 & 0.640 $\pm$ 0.178 \\
      GAT~\cite{55} & 93,095 & 7.03 & 0.777 $\pm$ 0.038 & 238,358 & 12.14 & 0.671 $\pm$ 0.013 & 33,315 & 15.13 & 0.755 $\pm$ 0.064 & 437,039 & 20.02 & 0.808 $\pm$ 0.099 \\
      GATv2~\cite{56} & 185,911 & 7.14 & 0.806 $\pm$ 0.042 & 476,454 & 11.98 & 0.662 $\pm$ 0.024 & 66,419 & 15.21 & 0.754 $\pm$ 0.057 & 873,663 & 20.37 & 0.888 $\pm$ 0.152 \\
      GraphGPS~\cite{57} & 197,727 & 8.33 & 0.480 $\pm$ 0.077 & 306,654 & 11.08 & 0.370 $\pm$ 0.023 & 152,811 & 97.48 & 0.645 $\pm$ 0.096 & 455,847 & 131.80 & 0.881 $\pm$ 0.197 \\
      SSGClassifier~\cite{59} & 23,607 & 4.70 & 0.721 $\pm$ 0.108 & 59,910 & 9.15 & 0.578 $\pm$ 0.082 & 8,611 & 8.95 & 0.762 $\pm$ 0.113 & 109,695 & 117.03 & 0.800 $\pm$ 0.211 \\
      Q-GCN~\cite{xiong2022pseudo} & 184,583 & 138.12 & 0.783 $\pm$ 0.008 & 475,014  & 132.90 & 0.663 $\pm$ 0.008 & 64,643 & 145.27 & 0.735 $\pm$ 0.009 & - & - & - \\
      TGNN~\cite{luo2024classic} & 1,899,783  & 7.97 & 0.857 $\pm$ 0.021 & 3,642,886  & 7.88 & 0.691 $\pm$ 0.035 & 1,182,211 & 18.31 & 0.848 $\pm$ 0.016 & - & - & - \\
      PR-CapsNet & 93,650  & 8.34 & \textbf{0.876 $\pm$ 0.012} & 238,910 & 8.52 & \textbf{0.755 $\pm$ 0.019} & 33,858 & 8.57 & \textbf{0.851 $\pm$ 0.004} & 437,618 & 13.47 & \textbf{0.915 $\pm$ 0.031} \\
      \hline
    \end{tabular}%
  }
  \caption{Performances on node classification benchmarks. The unit of Runtime is seconds.}
  \label{tab:node_classification_results_std_3dp}
\end{table*}

\section{Experiments}
\subsection{Experimental Settings}
\noindent\textbf{Datasets.} To thoroughly evaluate the performance and generalization capabilities of PR-CapsNet, we conducted experiments on multiple publicly available benchmarks, covering two fundamental graph analysis tasks: node classification and graph classification.
For the node classification task, we employed four widely recognized datasets: Cora~\cite{46}, Citeseer~\cite{47}, PubMed~\cite{48}, and CoauthorCS~\cite{49}. These datasets are standard benchmarks for evaluating models on tasks that predict individual node properties within a single large graph, commonly utilized in semi-supervised learning settings. 
For the graph classification task, we utilized five distinct datasets: MUTAG~\cite{50}, NCI1~\cite{51}, NCI109~\cite{51}, DD~\cite{52}, and PROTEINS~\cite{53}. These datasets span a variety of domains, including bioinformatics and chemistry, and exhibit diverse structural properties. This diversity allows for a robust assessment of PR-CapsNet's generalization ability across different graph data types. 

\noindent\textbf{Metrics.} We deploy Mean Accuracy (Acc) to assess the model's overall classification performance and predictive capability. We also report the parameter size and the evaluation runtime.

\noindent\textbf{Baselines.} For the graph classification task, we compare PR-CapsNet with DGCNN~\cite{54}, GAT~\cite{55}, GATv2~\cite{56}, GIN~\cite{57}, GraphGPS~\cite{58}, SSGClassifier~\cite{59}, Q-GCN~\cite{xiong2022pseudo}, and TGNN~\cite{luo2024classic}. 
For the node classification task, we compare PR-CapsNet with DGCNN~\cite{54}, GAT~\cite{55}, GATv2~\cite{56}, GraphGPS\cite{57}, RieGrace\cite{61}, SSGClassifier\cite{59}, GRDL~\cite{wang2024graph}, and GRATIN~\cite{abbahaddougraph}. 

\noindent\textbf{Implementation Details.} All training was performed on an NVIDIA RTX 3090. In the training stage, PR-CapsNet incorporates a GNN layer that produces 64-dimensional node embeddings. All 3 hidden capsule layers have 19 dimensions (comprising 9 spatial and $9+1$ temporal dimensions). An initial $\beta$ value of -1 was set. Model optimization was performed using the Riemannian Adam optimizer over 100 training epochs. A learning rate of 0.001 was employed throughout the training process. L2 weight decay of 0.0001 and a dropout rate of 0.5 are set to enhance robustness and mitigate potential overfitting. The dynamic routing mechanism, which governs the information flow and agreement between capsule layers, was configured to perform 3 iterations per routing step. The training was conducted using a batch size of 16. 

\subsection{Graph Classification}
Table \ref{tab:graph_classification_results_std_3dp} reveal that PR-CapsNet consistently demonstrates strong and superior performance, achieving the leading mean test accuracy on all 4 benchmark datasets. 
Specifically, PR-CapsNet achieves the highest mean accuracy on the PROTEINS dataset. This result exemplifies PR-CapsNet's capability to handle complex biological graph structures effectively and learn more discriminative representations. Furthermore, PR-CapsNet often exhibits competitive or lower standard deviations than baselines across the datasets, suggesting stable and reliable performance, particularly on the NCI1 dataset with its exceptionally low standard deviation.  Besides, the runtime and parameter size of PR-CapsNet achieve a reasonable cost compromise with the highest classification performance.
These results strongly validate the effectiveness of our PR-CapsNet in enhancing graph representation learning for complex graph structures.

\subsection{Node Classification}
Table \ref{tab:node_classification_results_std_3dp} reveals that PR-CapsNet consistently demonstrates strong and superior performance, achieving the leading mean test accuracy on all four benchmark datasets under consideration. This consistent performance across diverse datasets highlights PR-CapsNet's robustness and ability to effectively capture node-level discriminative features within complex and non-Euclidean graph structures. While Table \ref{tab:node_classification_results_std_3dp} provides a complete, detailed comparison, clearly showing PR-CapsNet's dominance across the board, we highlight one key aspect of its superiority. 
Specifically, on the Citeseer dataset, PR-CapsNet achieves the highest mean accuracy, demonstrating a significant improvement of approximately 6.4 percentage points over the following best-performing baseline TGNN. This result exemplifies PR-CapsNet's capability to learn representations for nodes in citation networks effectively. Furthermore, PR-CapsNet often exhibits competitive or lower standard deviations than baselines across the datasets, suggesting stable and reliable performance, particularly on the PubMed dataset with its exceptionally low standard deviation. 
These results strongly validate the effectiveness of our PR-CapsNet in enhancing graph representation learning for complex graph structures in the context of node classification.

\subsection{Qualitative Analysis}
\begin{figure}[t]
\centering
\resizebox{\linewidth}{!}{%
\begin{tabular}{ccc}
\textbf{DGCNN} & \textbf{GATv2} & \textbf{SSGClassifier} \\
\includegraphics[width=0.16\textwidth]{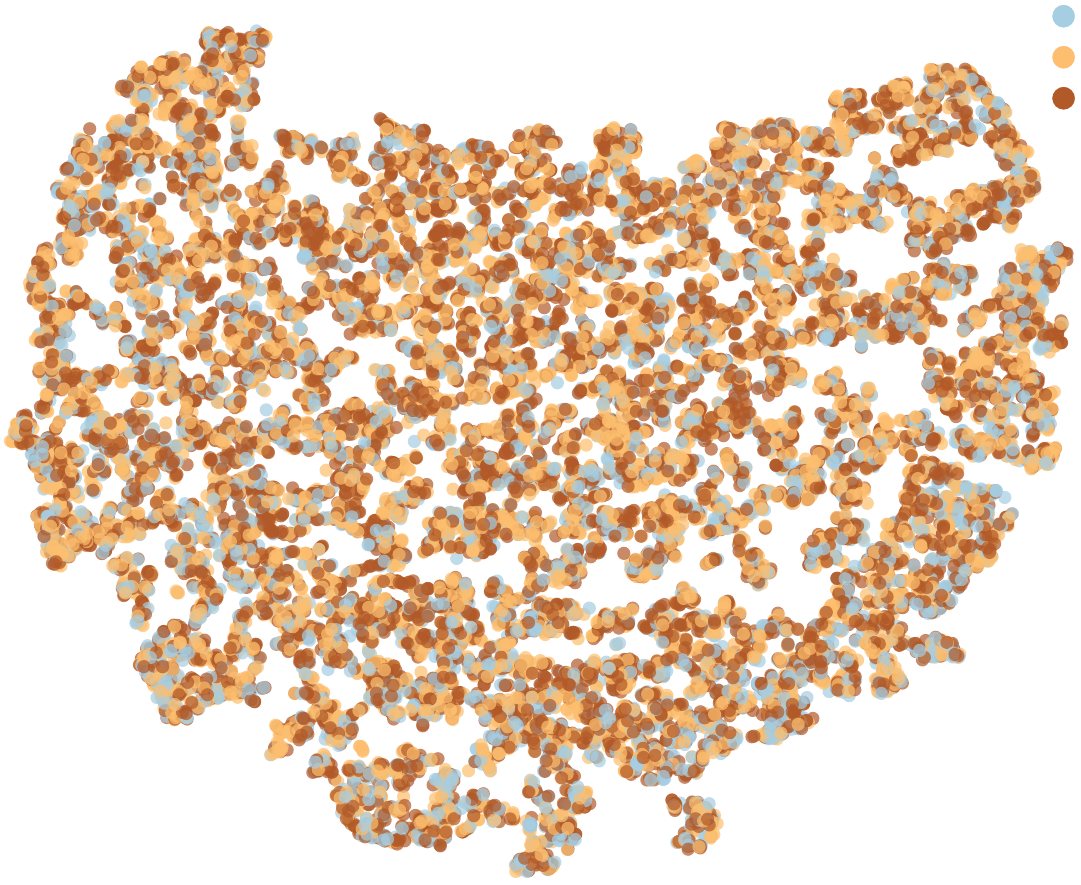} &
\includegraphics[width=0.16\textwidth]{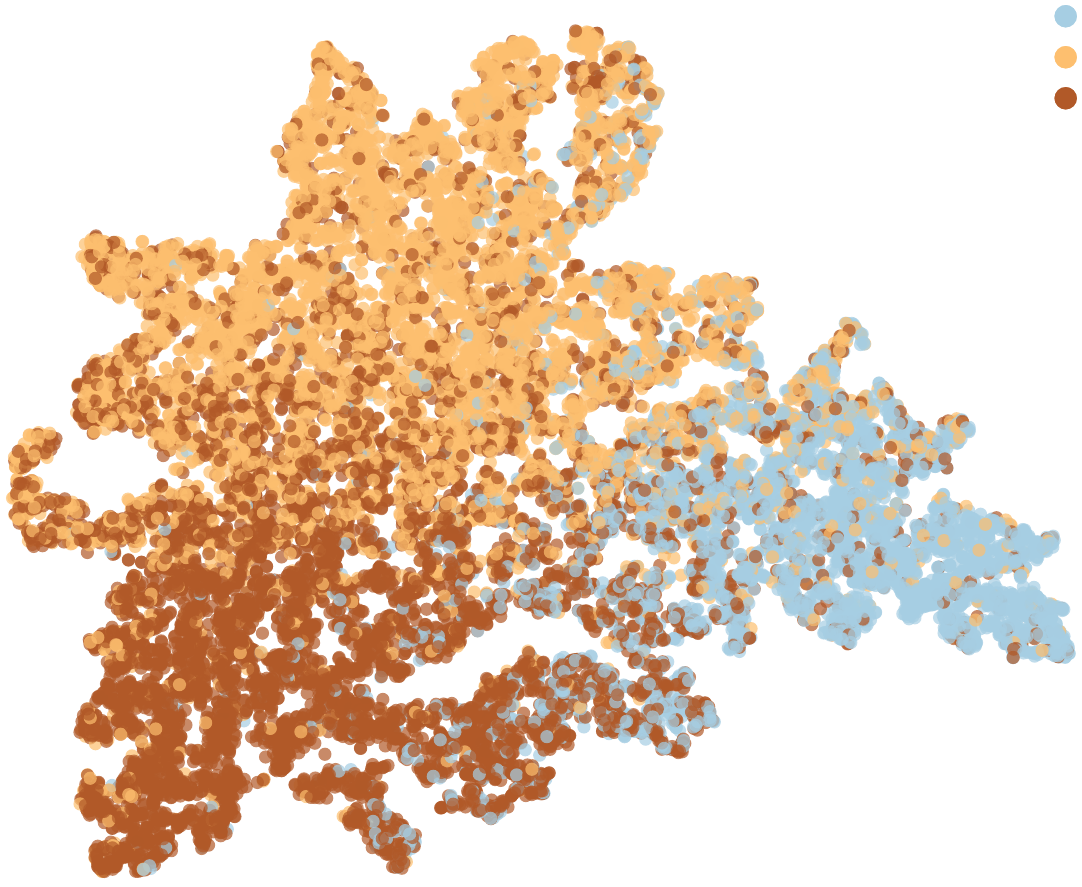}  &
\includegraphics[width=0.16\textwidth]{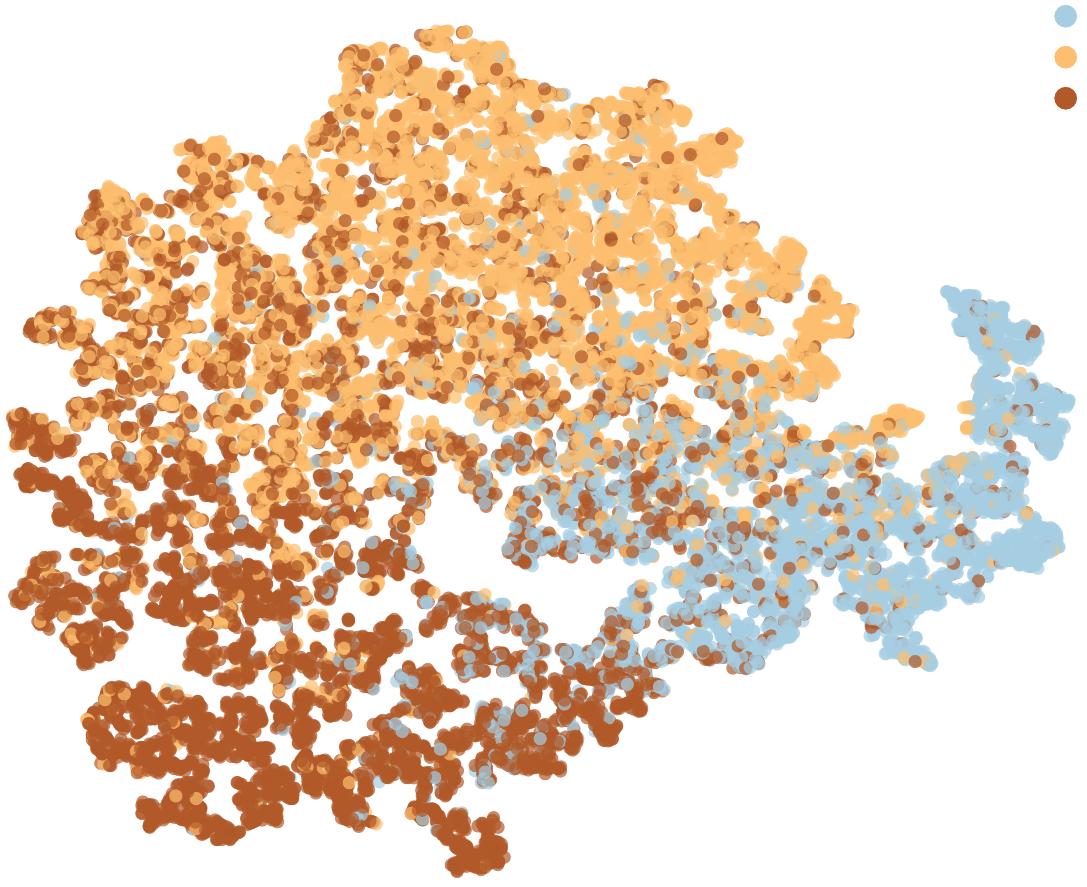} 
\\ 
\textbf{Q-GCN} & \textbf{TGNN} & \textbf{PR-CapsNet}  \\
\includegraphics[width=0.16\textwidth]{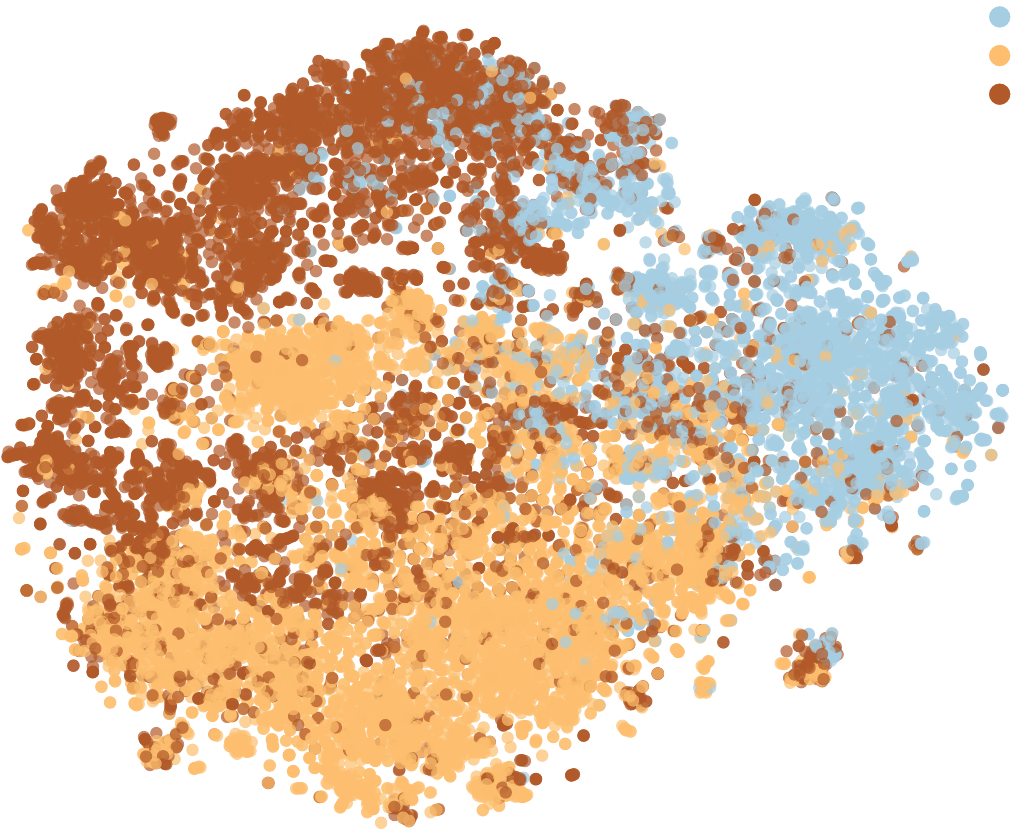} &
\includegraphics[width=0.16\textwidth]{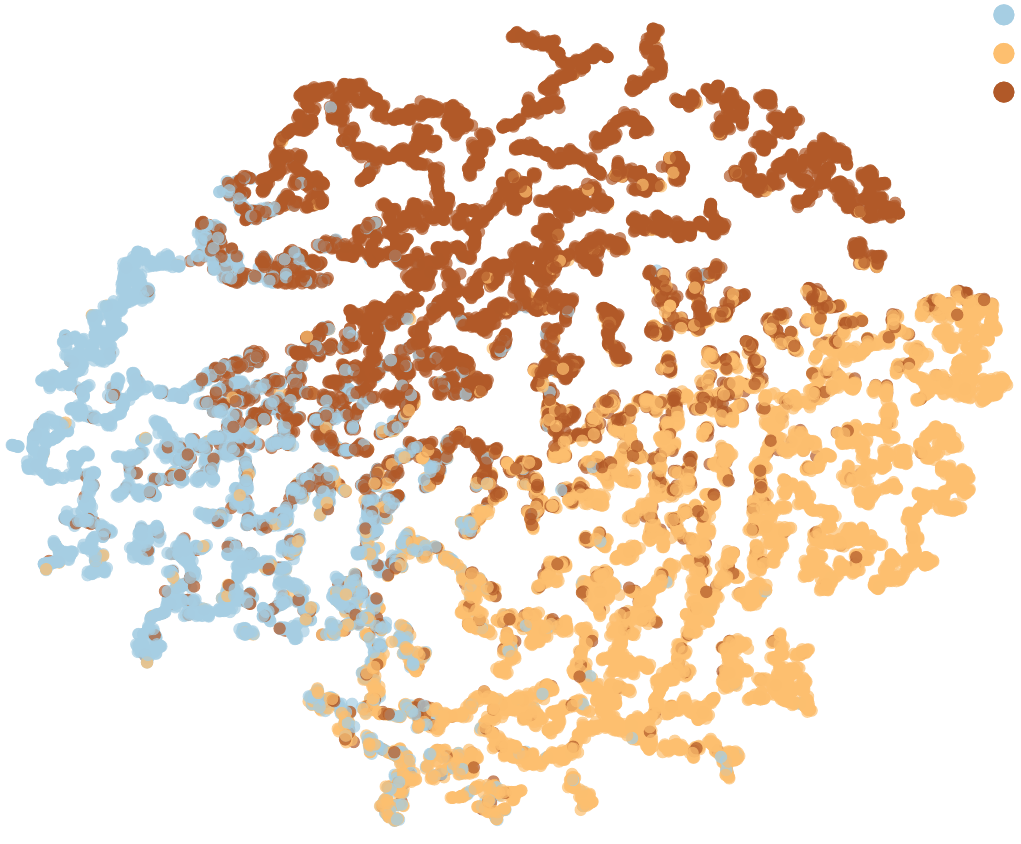} &
\includegraphics[width=0.16\textwidth]{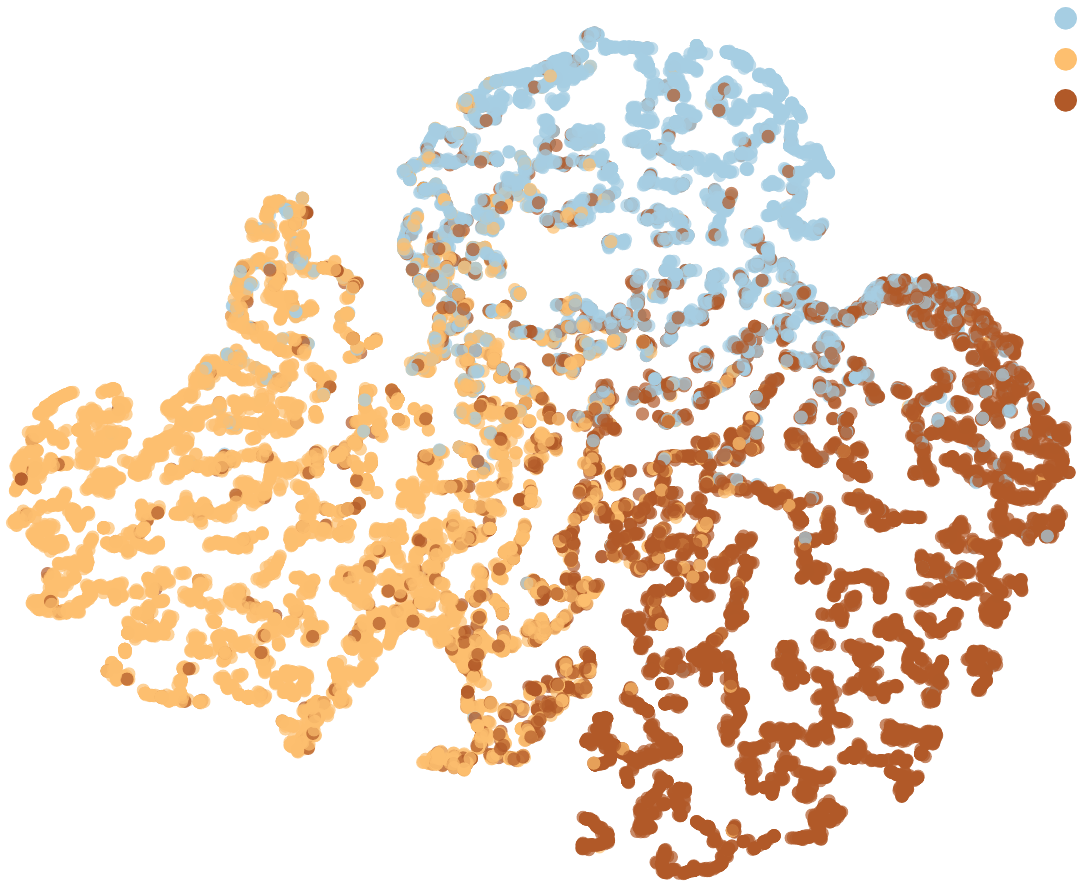} \\
\end{tabular}
}
\caption{t-SNE visualization of PubMed. Dark brown denotes the class of Experimental, light orange denotes Type 1, and light blue denotes Type 2.}
\label{tab:pubmed_tsne_visualizations_2x3}
\end{figure}

To better show the effect of our method in effectively capturing the rich information behind the graph using the idea of Capsules and the powerful spatial representation ability of pseudo-Riemannian tangent space, we use t-SNE to plot the graph distribution of different methods on the PubMed in Figure~\ref{tab:pubmed_tsne_visualizations_2x3}. 
The results reveal the varying efficacy of different methods in producing class-distinguishable node embeddings from the PubMed classification task. Baseline approaches, such as GATv2 and Q-GCN, yield distributions with notable inter-class overlap. DGCNN and SSGClassifier generally produce embeddings that lead to more discernible, though not entirely discrete, class clusters.  
Only the distribution resulting from TGNN and PR-CapsNet exhibits a more apparent separation of all three node classes than these other visualised methods, but the clusters corresponding to each class in PR-CapsNet appear more compact and exhibit less intermingling at their boundaries. This suggests that PR-CapsNet generates more class-discriminative representations effectively.

\begin{table}[t]
    \centering
    \resizebox{\linewidth}{!}{
    \begin{tabular}{|c|c|c|c|c|}
    \hline
    Pseudo-Riemannian & Capsnet & Cora Acc ($\uparrow$) & Pubmed Acc ($\uparrow$)\\
    \hline
    \cmark & \cmark & \textbf{0.838 $\pm$ 0.006} & \textbf{0.824 $\pm$ 0.010} \\
    \cmark & \xmark & $0.801 \pm 0.023$ & $0.803 \pm 0.004$ \\
    \xmark & \cmark & $0.781 \pm 0.006$ & $0.775 \pm 0.155$ \\
    \xmark & \xmark & $0.753 \pm 0.014$ & $0.747 \pm 0.021$ \\
    \hline
    \end{tabular}
    }
    \caption{Ablation Study of Pseudo-Riemannian Geometry and Capsule Components in PR-CapsNet for Node Classification on Cora and PubMed Datasets}
    \label{tab:Ablation study of PR-CapsNet}
\end{table}

\begin{table}[t]
\centering
\begin{tabular}{ccc} 
\toprule
{Routing Method} & {Classifier} & {Accuracy} \\
\midrule
ACR      & PRCC    & 0.755 \\
PCR      & PRCC    & 0.698 \\
Euclidean & PRCC   & 0.681 \\
ACR      & Linear  & 0.650 \\
\bottomrule
\end{tabular}
\caption{Accuracy comparison of routing methods and classifiers on CiteSeer.}
\label{tab:routing_accuracy}
\end{table}

\subsection{Ablation study}
\noindent\textbf{1) Ablation study of components in PR-CapsNet.}
We evaluated component contributions of PR-CapsNet variants like Pseudo-Riemannian-based GNNs (PRGNN), CapsNet, and GNNs via ablation studies on node classification tasks. 
The results in Table \ref{tab:Ablation study of PR-CapsNet} indicate that the progressive integration of pseudo-Riemannian geometry and capsule networks yields significant performance improvements in graph modelling tasks. For instance, on the Cora dataset, the complete PR-CapsNet model achieves an accuracy of 0.838, representing a 3.7$\%$ absolute improvement over the PRGNN and a 5.7$\%$ increase compared to the capsule network alone (0.781). This demonstrates that the synergistic combination of geometric structure encoding and hierarchical feature abstraction through dynamic routing creates a multiplicative effect rather than additive gains. Similarly, on PubMed, the PR-CapsNet model outperforms both individual components by 2.1$\%$ and 4.9$\%$, respectively, further validating the complementary nature of these architectural elements. The consistent performance advantage across datasets highlights the value of integrating geometric priors with adaptive learning mechanisms for complex graph representation tasks. In addition, we also investigate the effects of different routing methods (Euclidean from original CapsNet, PCR,  ACR) and classifiers (PRCC and simple Linear layer). The experiment results on CiteSeer are shown in Table~\ref{tab:routing_accuracy}. We can see that all components, PCR, ACR, and PRCC, can improve the classification accuracy, showing the effectiveness of our proposed improvements.

\begin{figure}[t]
    \centering
    \includegraphics[width=0.95\linewidth]{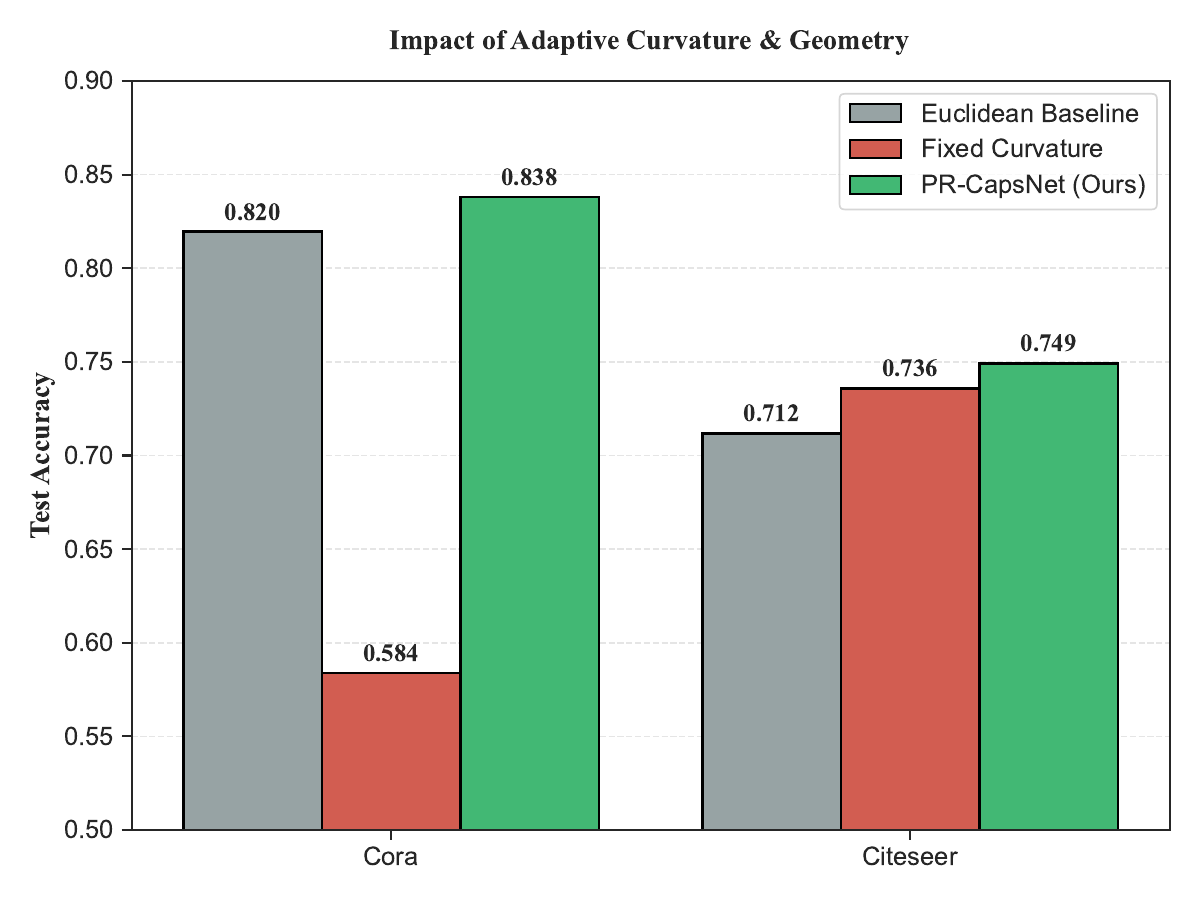}
    \caption{Ablation study of geometric components on Cora and CiteSeer. ``Fixed Curvature'' denotes a model with a frozen pseudo-Riemannian metric.}
    \label{fig:ablation_component}
\end{figure}

We also conducted an ablation study comparing PR-CapsNet against a baseline using a fixed pseudo-Riemannian metric (``Fixed Curvature'') to validate the necessity of our Adaptive Curvature Routing (ACR). As shown in Figure~\ref{fig:ablation_component}, the fixed curvature model suffers a drastic performance collapse on Cora, whereas our adaptive approach achieves 83.8\%. This confirms that real-world graphs exhibit complex local topologies that cannot be adequately modeled by a single fixed-curvature manifold.

\noindent\textbf{2) Analysis of Routing Iterations.}
In our PR-CapsNet, routing iteration dynamically refines connections between capsules, enabling hierarchical feature binding through agreement. We investigate the impact of different numbers of routing iterations.
\begin{figure}[h] 
    \centering    
    \includegraphics[width=\linewidth]{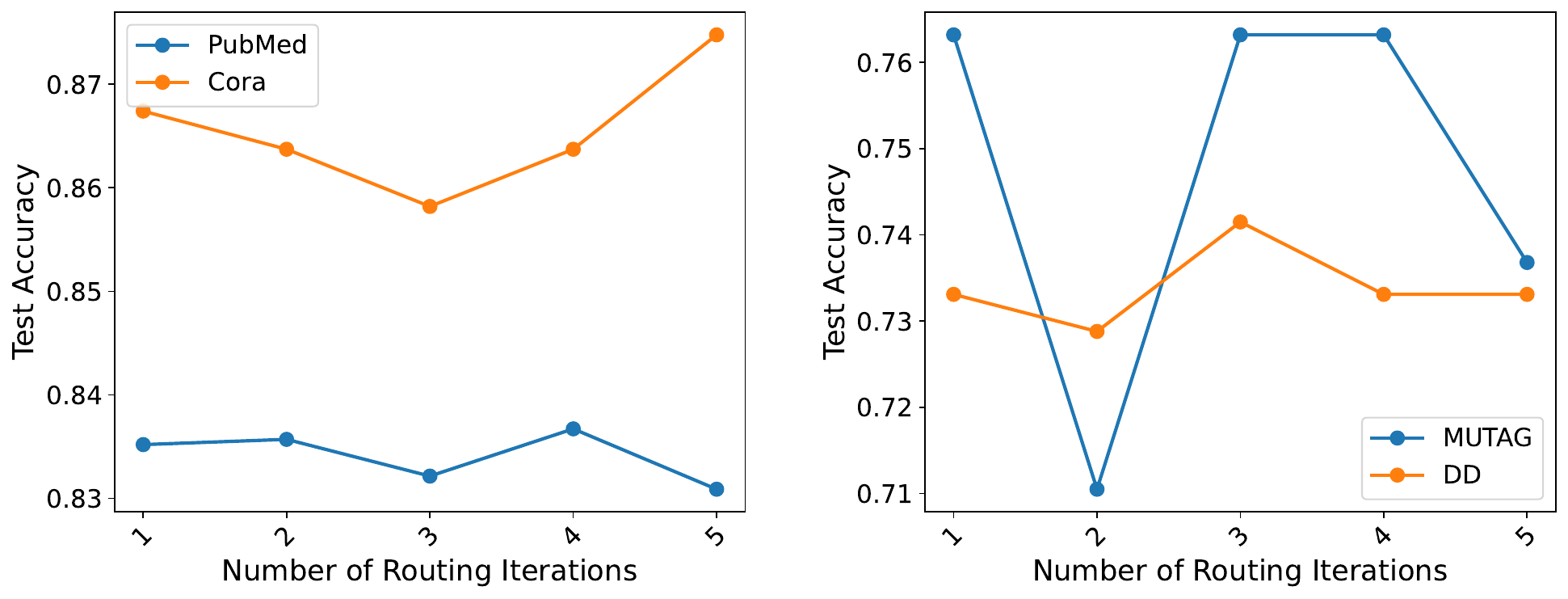} %
    \caption{Effects of the number of routing iterations.} 
    \label{fig:routing_iterations_analysis} 
\end{figure}
The results in Figure~\ref{fig:routing_iterations_analysis} indicate that performance generally improves with more iterations up to a point, but does not always increase monotonically thereafter. 
This non-monotonic pattern confirms the existence of an optimal iteration point in dynamic routing: insufficient iterations fail to exploit capsule interactions in pseudo-Riemannian space fully. In contrast, excessive iterations disrupt geometric coherence through gradient saturation.

\begin{figure}[t] 
    \centering 
    \includegraphics[width=\linewidth]{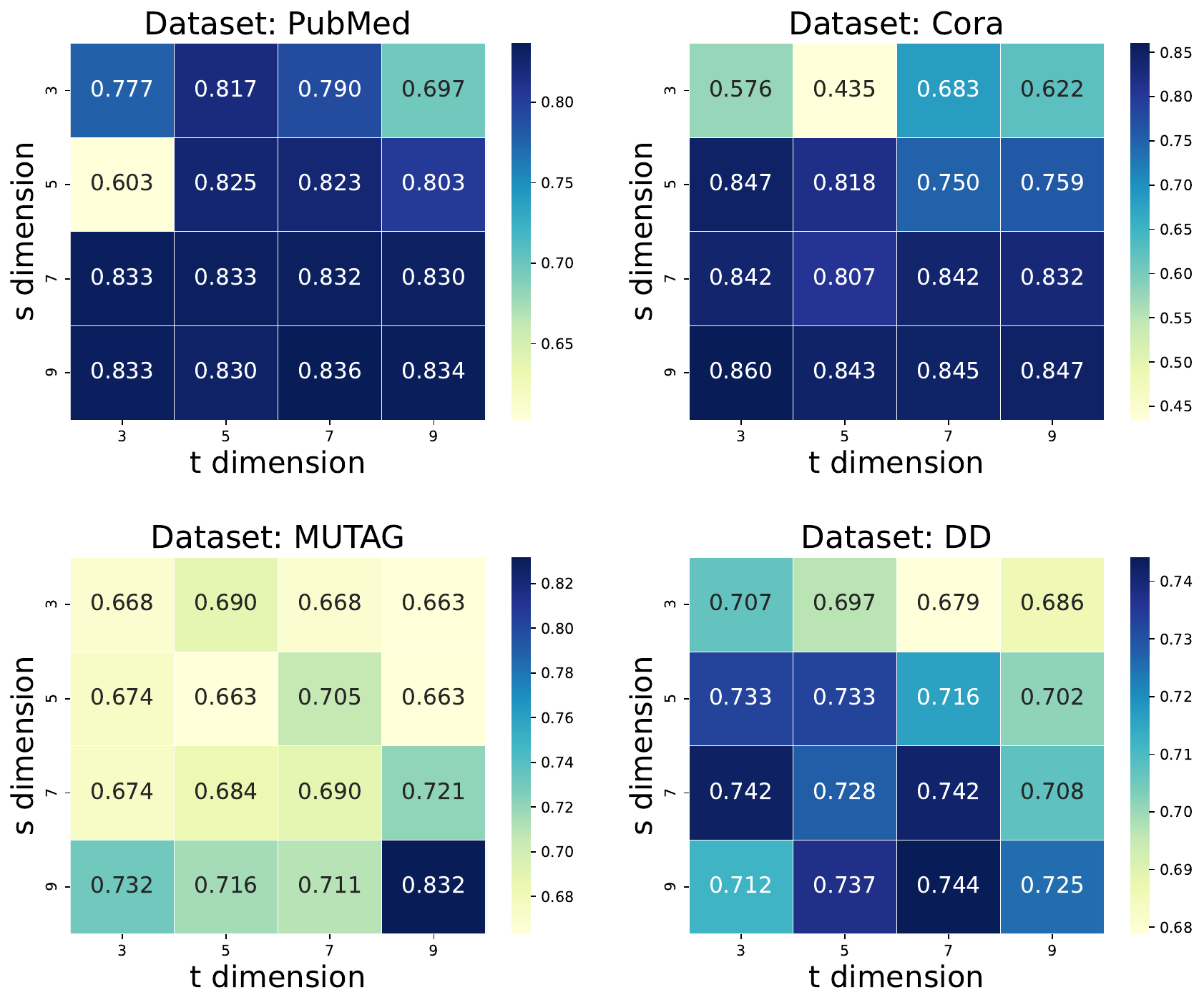} 
    \caption{Effects of different spatiotemporal dimensions on test accuracy.} 
    \label{fig:spatiotemporal_dimensions_analysis} 
\end{figure} 

\noindent\textbf{3) Analysis of spatiotemporal dimensions.}
In our PR-CapsNet, spatiotemporal dimensions jointly form a pseudo-Riemannian space representing capsule entities. Spatial dimensions capture Euclidean geometric structures. Temporal dimensions utilise indefinite metric properties, capturing non-Euclidean structures such as hierarchy and causality. Together, these dimensions define the geometric environment for data embedding. Thus, we investigate the impact of different spatial and temporal dimensions. The results are shown in Figure \ref{fig:spatiotemporal_dimensions_analysis}. The observed trends indicate that increasing spatial dimensions generally leads to significant improvements in model performance. In contrast, the benefits of growing temporal dimensions become negligible at high spatial dimensions. These findings suggest the need for flexible adjustment of spatiotemporal dimension configurations based on the specific characteristics of the dataset. The PR-CapsNet method, by integrating both spatial and temporal dimensions, effectively adapts to the complex structures of diverse datasets, thereby enhancing the model's representational capability and generalization performance.
With the increase of the $s$ dimension from 3 to 7 and 9, there is a notable improvement in model performance. When the $t$ dimension is 3, the accuracy improves from 0.697 to over 0.830. This trend indicates that higher spatial dimensions are crucial for enhancing model performance. Moreover, at high $s$ dimensions (7 and 9), variations in the $t$ dimension have minimal effects on accuracy, with scores consistently above 0.830 across different $t$ dimensions. This suggests that at sufficiently high spatial dimensions, increasing the $t$ dimension yields limited gains in performance.
Besides, the relationship between spatial dimensions and model performance within the PubMed dataset exhibits non-linear characteristics. While an increase in the temporal dimension can improve performance at lower spatial dimensions, this effect diminishes significantly at higher spatial dimensions. This observation underscores the fundamental and decisive role of spatial dimensions in determining model performance. In contrast, the role of temporal dimensions appears more supportive and contingent upon the configuration of spatial dimensions.
Overall, trends observed in the PubMed dataset align with those seen in other datasets, indicating that increasing spatial dimensions generally leads to significant improvements in model performance. In contrast, the benefits of growing temporal dimensions become negligible at high spatial dimensions. These findings suggest the need for flexible adjustment of spatiotemporal dimension configurations based on the specific characteristics of the dataset. The PR-CapsNet method, by integrating both spatial and temporal dimensions, effectively adapts to the complex structures of diverse datasets, thereby enhancing the model's representational capability and generalization performance.
			
\begin{figure}[t]
    \centering
    \begin{minipage}{0.48\linewidth}
        \centering
        \includegraphics[width=\linewidth]{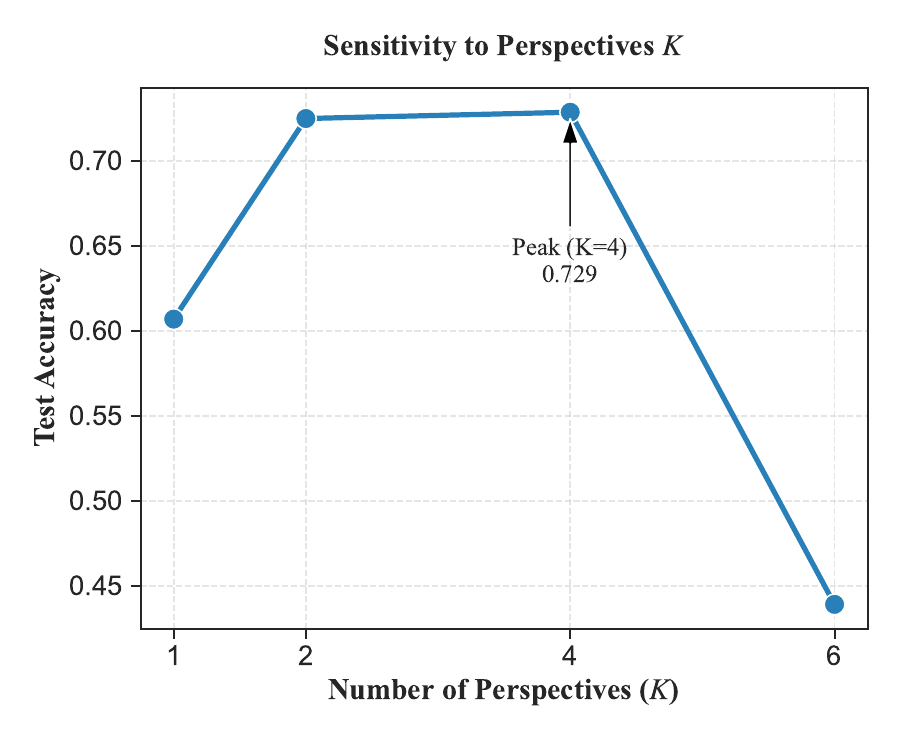}
        \caption{Sensitivity analysis of perspectives $K$ on Cora. Performance peaks at $K=4$.}
        \label{fig:k_sensitivity}
    \end{minipage}
    \hfill
    \begin{minipage}{0.48\linewidth}
        \centering
        \includegraphics[width=\linewidth]{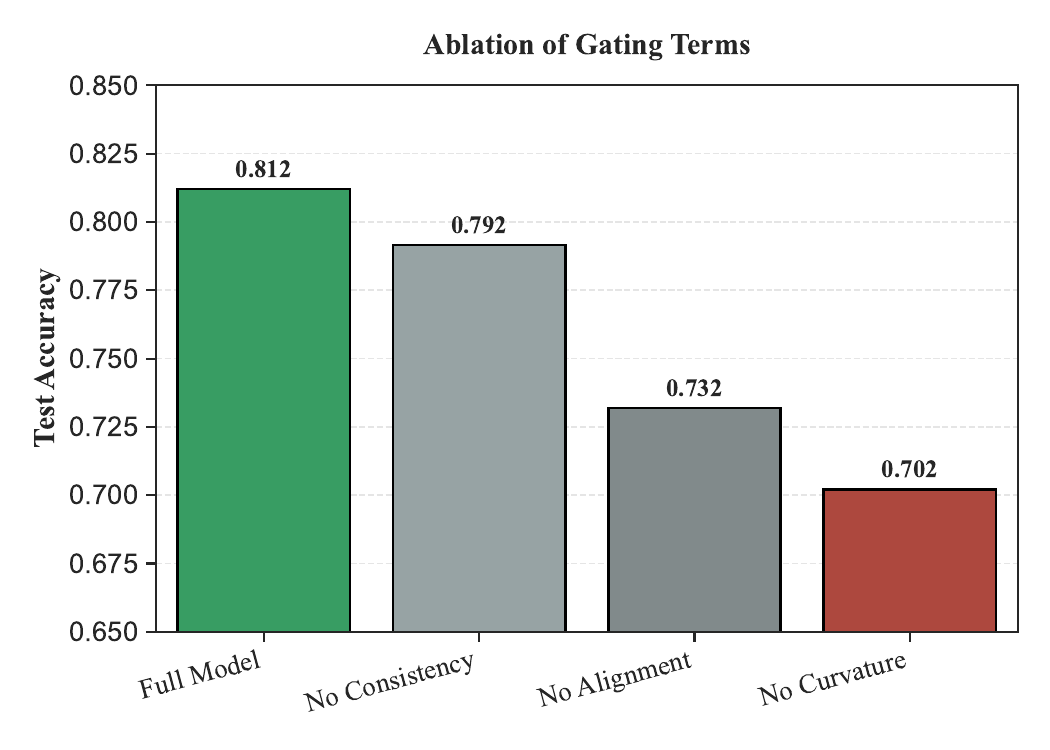}
        \caption{Ablation of gating terms on PubMed. Curvature ($C$) is the most critical component.}
        \label{fig:gating_ablation}
    \end{minipage}
\end{figure}

We further analyzed the internal mechanisms of ACR. Figure~\ref{fig:gating_ablation} presents a fine-grained ablation of the gating terms (Eq.~\ref{eq:gating_weights_new}) on PubMed. Removing the \textit{Curvature Compatibility} ($C$) term results in the largest performance drop, confirming that matching local manifold curvature is the most critical factor. Regarding the number of perspectives $K$, Figure~\ref{fig:k_sensitivity} shows that accuracy improves as $K$ increases to 4, balancing structural diversity and model complexity. Increasing $K$ further leads to overfitting, justifying our default setting of $K=4$.

\section{Conclusion}
In this paper, we extend the Euclidean capsule routing into geodesically disconnected pseudo-Riemannian manifolds and derive a PR-CapsNet to model data in pseudo-Riemannian manifolds of adaptive curvature in the context of graph representation learning. Extensive experiments on node and graph classification benchmarks validate the superiority of PR-CapsNet for modeling complex graph structures. Our PR-CapsNet provides a powerful solution for learning complex graph structures, facilitating further research in geometric deep learning.

\section{Acknowledgement}
This work was supported in part by the National Natural Science Foundation of China (NSFC) under Grant No. 62206314, GuangDong Basic and Applied Basic Research Foundation under Grant No. 2022A1515011835, Science and Technology Projects in Guangzhou under Grant No. 2024A04J4388.

\section{Ethical Considerations}
This paper investigates graph learning via a novel Pseudo-Riemannian Capsule Network with Adaptive Curvature Routing. We extend the Euclidean capsule routing into geodesically disconnected pseudo-Riemannian manifolds to overcome the limitations of the Euclidean space-based CapsNet on modeling complex graphs with diverse local structures, like hierarchical structure, cluster structure, cyclic structure, etc. As an improvement of CapsNet, our PR-CapsNet can be generalized to other fields that involve graph structure, such as social networks, recommendation systems, financial risk control, bioinformatics, transportation planning, and knowledge graphs. As a deep learning algorithm, it may produce biased results and poor performance if trained on small-scale data. Therefore, we should ensure the data volume when applying our PR-CapsNet on specific tasks.    

\bibliographystyle{unsrt}
\balance
\bibliography{ref}

\begin{thebibliography}{10}

\bibitem{26}
Thomas~N Kipf and Max Welling.
\newblock Semi-supervised classification with graph convolutional networks.
\newblock {\em arXiv preprint arXiv:1609.02907}, 2016.

\bibitem{velickovic2018graph}
Petar Velickovic, Guillem Cucurull, Arantxa Casanova, Adriana Romero, Pietro
  Liò, and Yoshua Bengio.
\newblock Graph attention networks.
\newblock In {\em 6th International Conference on Learning Representations,
  ICLR 2018, Vancouver, BC, Canada, April 30 - May 3, 2018, Conference Track
  Proceedings}. OpenReview.net, 2018.

\bibitem{wu2019simplifying}
Felix Wu, Amauri H~Souza Jr, Tianyi Zhang, Christopher Fifty, Tao Yu, and
  Kilian~Q Weinberger.
\newblock Simplifying graph convolutional networks.
\newblock In {\em Proceedings of the 36th International Conference on Machine
  Learning, ICML 2019, 9-15 June 2019, Long Beach, California, USA}, volume~97
  of {\em Proceedings of Machine Learning Research}, pages 6861--6871. PMLR,
  2019.

\bibitem{bruna2014spectral}
Joan Bruna, Wojciech Zaremba, Arthur Szlam, and Yann LeCun.
\newblock Spectral networks and locally connected networks on graphs.
\newblock In Yoshua Bengio and Yann LeCun, editors, {\em 2nd International
  Conference on Learning Representations, ICLR 2014, Banff, AB, Canada, April
  14-16, 2014, Conference Track Proceedings}, 2014.

\bibitem{58}
Ladislav Ramp{\'a}{\v{s}}ek, Michael Galkin, Vijay~Prakash Dwivedi, Anh~Tuan
  Luu, Guy Wolf, and Dominique Beaini.
\newblock Recipe for a general, powerful, scalable graph transformer.
\newblock {\em Advances in Neural Information Processing Systems},
  35:14501--14515, 2022.

\bibitem{boguna2021network}
Marián Boguñá, Ivan Bonamassa, Manlio~De Domenico, Shlomo Havlin, Dmitri
  Krioukov, and M~Ángeles Serrano.
\newblock Network geometry.
\newblock {\em Nature Reviews Physics}, pages 1--22, 2021.

\bibitem{gu2019learning}
Albert Gu, Frederic Sala, Beliz Gunel, and Christopher Ré.
\newblock Learning mixed-curvature representations in product spaces.
\newblock In {\em 7th International Conference on Learning Representations,
  ICLR 2019, New Orleans, LA, USA, May 6-9, 2019}. OpenReview.net, 2019.

\bibitem{45}
Sara Sabour, Nicholas Frosst, and Geoffrey~E Hinton.
\newblock Dynamic routing between capsules.
\newblock {\em Advances in neural information processing systems}, 30, 2017.

\bibitem{28}
Zhang Xinyi and Lihui Chen.
\newblock Capsule graph neural network.
\newblock In {\em International conference on learning representations}, 2018.

\bibitem{29}
Yang Li, Wei Zhao, Erik Cambria, Suhang Wang, and Steffen Eger.
\newblock Graph routing between capsules.
\newblock {\em Neural Networks}, 143:345--354, 2021.

\bibitem{41}
Weize Chen, Xu~Han, Yankai Lin, Hexu Zhao, Zhiyuan Liu, Peng Li, Maosong Sun,
  and Jie Zhou.
\newblock Fully hyperbolic neural networks.
\newblock {\em arXiv preprint arXiv:2105.14686}, 2021.

\bibitem{42}
Anne Leyrat-Maurin and Dominique Barthes-Biesel.
\newblock Motion of a deformable capsule through a hyperbolic constriction.
\newblock {\em Journal of fluid mechanics}, 279:135--163, 1994.

\bibitem{krioukov2010hyperbolic}
Dmitri Krioukov, Fragkiskos Papadopoulos, Maksim Kitsak, Amin Vahdat, and
  Marián Boguná.
\newblock Hyperbolic geometry of complex networks.
\newblock {\em Physical Review E}, 82(3):036106, 2010.

\bibitem{6}
Maximillian Nickel and Douwe Kiela.
\newblock Poincar{\'e} embeddings for learning hierarchical representations.
\newblock {\em Advances in neural information processing systems}, 30, 2017.

\bibitem{wilson2014spherical}
Richard~C Wilson, Edwin~R Hancock, Elżbieta Pekalska, and Robert~PW Duin.
\newblock Spherical and hyperbolic embeddings of data.
\newblock {\em IEEE transactions on pattern analysis and machine intelligence},
  36(11):2255--2269, 2014.

\bibitem{meng2019spherical}
Yu~Meng, Jiaxin Huang, Guangyuan Wang, Chao Zhang, Honglei Zhuang, Lance~M
  Kaplan, and Jiawei Han.
\newblock Spherical text embedding.
\newblock In {\em Advances in Neural Information Processing Systems 32: Annual
  Conference on Neural Information Processing Systems 2019, NeurIPS 2019,
  December 8-14, 2019, Vancouver, BC, Canada}, pages 8206--8215, 2019.

\bibitem{defferrard2019deepsphere}
Michaël Defferrard, Nathanaël Perraudin, Tomasz Kacprzak, and Raphael Sgier.
\newblock Deepsphere: towards an equivariant graph-based spherical cnn.
\newblock In {\em ICLR 2019 Workshop on Representation Learning on Graphs and
  Manifolds}, 2019.

\bibitem{davidson2018hyperspherical}
Tim~R Davidson, Luca Falorsi, Nicola~De Cao, Thomas Kipf, and Jakub~M Tomczak.
\newblock Hyperspherical variational auto-encoders.
\newblock In {\em Proceedings of the Thirty-Fourth Conference on Uncertainty in
  Artificial Intelligence, UAI 2018, Monterey, California, USA, August 6-10,
  2018}, pages 856--865. AUAI Press, 2018.

\bibitem{laub2004feature}
Julian Laub and Klaus-Robert Müller.
\newblock Feature discovery in non-metric pairwise data.
\newblock {\em The Journal of Machine Learning Research}, 5:801--818, 2004.

\bibitem{oneill1983semi}
Barrett O'neill.
\newblock {\em Semi-riemannian geometry with applications to relativity},
  volume 103.
\newblock Academic pres, 1983.

\bibitem{15}
Marc Law and Jos Stam.
\newblock Ultrahyperbolic representation learning.
\newblock {\em Advances in neural information processing systems},
  33:1668--1678, 2020.

\bibitem{13}
Bo~Xiong, Shichao Zhu, Nico Potyka, Shirui Pan, Chuan Zhou, and Steffen Staab.
\newblock Pseudo-riemannian graph convolutional networks.
\newblock {\em Advances in Neural Information Processing Systems},
  35:3488--3501, 2022.

\bibitem{1}
Maximilian Nickel, Volker Tresp, Hans-Peter Kriegel, et~al.
\newblock A three-way model for collective learning on multi-relational data.
\newblock In {\em Icml}, volume~11, pages 3104482--3104584, 2011.

\bibitem{2}
Antoine Bordes, Nicolas Usunier, Alberto Garcia-Duran, Jason Weston, and Oksana
  Yakhnenko.
\newblock Translating embeddings for modeling multi-relational data.
\newblock {\em Advances in neural information processing systems}, 26, 2013.

\bibitem{3}
Frederic Sala, Chris De~Sa, Albert Gu, and Christopher R{\'e}.
\newblock Representation tradeoffs for hyperbolic embeddings.
\newblock In {\em International conference on machine learning}, pages
  4460--4469. PMLR, 2018.

\bibitem{4}
Mikhael Gromov, Misha Katz, Pierre Pansu, and Stephen Semmes.
\newblock {\em Metric structures for Riemannian and non-Riemannian spaces},
  volume 152.
\newblock Springer, 1999.

\bibitem{5}
Benjamin~Paul Chamberlain, James Clough, and Marc~Peter Deisenroth.
\newblock Neural embeddings of graphs in hyperbolic space.
\newblock {\em arXiv preprint arXiv:1705.10359}, 2017.

\bibitem{7}
Octavian Ganea, Gary B{\'e}cigneul, and Thomas Hofmann.
\newblock Hyperbolic neural networks.
\newblock {\em Advances in neural information processing systems}, 31, 2018.

\bibitem{8}
Ines Chami, Zhitao Ying, Christopher R{\'e}, and Jure Leskovec.
\newblock Hyperbolic graph convolutional neural networks.
\newblock {\em Advances in neural information processing systems}, 32, 2019.

\bibitem{9}
Qi~Liu, Maximilian Nickel, and Douwe Kiela.
\newblock Hyperbolic graph neural networks.
\newblock {\em Advances in neural information processing systems}, 32, 2019.

\bibitem{10}
Shichao Zhu, Shirui Pan, Chuan Zhou, Jia Wu, Yanan Cao, and Bin Wang.
\newblock Graph geometry interaction learning.
\newblock {\em Advances in Neural Information Processing Systems},
  33:7548--7558, 2020.

\bibitem{11}
Yiding Zhang, Xiao Wang, Chuan Shi, Nian Liu, and Guojie Song.
\newblock Lorentzian graph convolutional networks.
\newblock In {\em Proceedings of the web conference 2021}, pages 1249--1261,
  2021.

\bibitem{12}
Jindou Dai, Yuwei Wu, Zhi Gao, and Yunde Jia.
\newblock A hyperbolic-to-hyperbolic graph convolutional network.
\newblock In {\em Proceedings of the IEEE/CVF conference on computer vision and
  pattern recognition}, pages 154--163, 2021.

\bibitem{14}
Aaron Sim, Maciej~L Wiatrak, Angus Brayne, P{\'a}id{\'\i} Creed, and Saee
  Paliwal.
\newblock Directed graph embeddings in pseudo-riemannian manifolds.
\newblock In {\em International Conference on Machine Learning}, pages
  9681--9690. PMLR, 2021.

\bibitem{16}
Bo~Xiong, Shichao Zhu, Mojtaba Nayyeri, Chengjin Xu, Shirui Pan, Chuan Zhou,
  and Steffen Staab.
\newblock Ultrahyperbolic knowledge graph embeddings.
\newblock In {\em Proceedings of the 28th ACM SIGKDD Conference on Knowledge
  Discovery and Data Mining}, pages 2130--2139, 2022.

\bibitem{17}
Yann LeCun, L{\'e}on Bottou, Yoshua Bengio, and Patrick Haffner.
\newblock Gradient-based learning applied to document recognition.
\newblock {\em Proceedings of the IEEE}, 86(11):2278--2324, 1998.

\bibitem{18}
Alex Krizhevsky, Ilya Sutskever, and Geoffrey~E Hinton.
\newblock Imagenet classification with deep convolutional neural networks.
\newblock {\em Advances in neural information processing systems}, 25, 2012.

\bibitem{19}
Kaiming He, Xiangyu Zhang, Shaoqing Ren, and Jian Sun.
\newblock Deep residual learning for image recognition.
\newblock In {\em Proceedings of the IEEE conference on computer vision and
  pattern recognition}, pages 770--778, 2016.

\bibitem{20}
Marco Gori, Gabriele Monfardini, and Franco Scarselli.
\newblock A new model for learning in graph domains.
\newblock In {\em Proceedings. 2005 IEEE international joint conference on
  neural networks, 2005.}, volume~2, pages 729--734. IEEE, 2005.

\bibitem{21}
Franco Scarselli, Marco Gori, Ah~Chung Tsoi, Markus Hagenbuchner, and Gabriele
  Monfardini.
\newblock The graph neural network model.
\newblock {\em IEEE transactions on neural networks}, 20(1):61--80, 2008.

\bibitem{22}
William~L Hamilton, Rex Ying, and Jure Leskovec.
\newblock Representation learning on graphs: Methods and applications.
\newblock {\em arXiv preprint arXiv:1709.05584}, 2017.

\bibitem{23}
Jie Zhou, Ganqu Cui, Shengding Hu, Zhengyan Zhang, Cheng Yang, Zhiyuan Liu,
  Lifeng Wang, Changcheng Li, and Maosong Sun.
\newblock Graph neural networks: A review of methods and applications.
\newblock {\em AI open}, 1:57--81, 2020.

\bibitem{24}
Alessio Micheli.
\newblock Neural network for graphs: A contextual constructive approach.
\newblock {\em IEEE Transactions on Neural Networks}, 20(3):498--511, 2009.

\bibitem{25}
Mathias Niepert, Mohamed Ahmed, and Konstantin Kutzkov.
\newblock Learning convolutional neural networks for graphs.
\newblock In {\em International conference on machine learning}, pages
  2014--2023. PMLR, 2016.

\bibitem{27}
Jindong Gu.
\newblock Interpretable graph capsule networks for object recognition.
\newblock In {\em Proceedings of the AAAI Conference on Artificial
  Intelligence}, volume~35, pages 1469--1477, 2021.

\bibitem{30}
Qingxing Cao, Wentao Wan, Keze Wang, Xiaodan Liang, and Liang Lin.
\newblock Linguistically routing capsule network for out-of-distribution visual
  question answering.
\newblock In {\em Proceedings of the IEEE/CVF International Conference on
  Computer Vision}, pages 1614--1623, 2021.

\bibitem{31}
Chenliang Li, Cong Quan, Li~Peng, Yunwei Qi, Yuming Deng, and Libing Wu.
\newblock A capsule network for recommendation and explaining what you like and
  dislike.
\newblock In {\em Proceedings of the 42nd international ACM SIGIR conference on
  research and development in information retrieval}, pages 275--284, 2019.

\bibitem{32}
Huan Lin, Fandong Meng, Jinsong Su, Yongjing Yin, Zhengyuan Yang, Yubin Ge, Jie
  Zhou, and Jiebo Luo.
\newblock Dynamic context-guided capsule network for multimodal machine
  translation.
\newblock In {\em Proceedings of the 28th ACM international conference on
  multimedia}, pages 1320--1329, 2020.

\bibitem{33}
Parnian Afshar, Shahin Heidarian, Farnoosh Naderkhani, Anastasia Oikonomou,
  Konstantinos~N Plataniotis, and Arash Mohammadi.
\newblock Covid-caps: A capsule network-based framework for identification of
  covid-19 cases from x-ray images.
\newblock {\em Pattern Recognition Letters}, 138:638--643, 2020.

\bibitem{34}
Parnian Afshar, Konstantinos~N Plataniotis, and Arash Mohammadi.
\newblock Capsule networks for brain tumor classification based on mri images
  and coarse tumor boundaries.
\newblock In {\em ICASSP 2019-2019 IEEE international conference on acoustics,
  speech and signal processing (ICASSP)}, pages 1368--1372. IEEE, 2019.

\bibitem{35}
Fudong Li, Xingyu Lu, and Jianjun Yuan.
\newblock Mha-corocapsule: multi-head attention routing-based capsule network
  for covid-19 chest x-ray image classification.
\newblock {\em IEEE Transactions on Medical Imaging}, 41(5):1208--1218, 2021.

\bibitem{36}
Karim Ahmed and Lorenzo Torresani.
\newblock Star-caps: Capsule networks with straight-through attentive routing.
\newblock {\em Advances in neural information processing systems}, 32, 2019.

\bibitem{37}
Fabio De~Sousa~Ribeiro, Georgios Leontidis, and Stefanos Kollias.
\newblock Introducing routing uncertainty in capsule networks.
\newblock {\em Advances in neural information processing systems},
  33:6490--6502, 2020.

\bibitem{38}
Miles Everett, Mingjun Zhong, and Georgios Leontidis.
\newblock Protocaps: A fast and non-iterative capsule network routing method.
\newblock {\em arXiv preprint arXiv:2307.09944}, 2023.

\bibitem{39}
Wei Zhao, Jianbo Ye, Min Yang, Zeyang Lei, Suofei Zhang, and Zhou Zhao.
\newblock Investigating capsule networks with dynamic routing for text
  classification.
\newblock {\em arXiv preprint arXiv:1804.00538}, 2018.

\bibitem{40}
Chunning Du, Haifeng Sun, Jingyu Wang, Qi~Qi, Jianxin Liao, Chun Wang, and Bing
  Ma.
\newblock Investigating capsule network and semantic feature on hyperplanes for
  text classification.
\newblock In {\em Proceedings of the 2019 Conference on Empirical Methods in
  Natural Language Processing and the 9th International Joint Conference on
  Natural Language Processing (EMNLP-IJCNLP)}, pages 456--465, 2019.

\bibitem{46}
Andrew~Kachites McCallum, Kamal Nigam, Jason Rennie, and Kristie Seymore.
\newblock Automating the construction of internet portals with machine
  learning.
\newblock {\em Information Retrieval}, 3:127--163, 2000.

\bibitem{47}
C~Lee Giles, Kurt~D Bollacker, and Steve Lawrence.
\newblock Citeseer: An automatic citation indexing system.
\newblock In {\em Proceedings of the third ACM conference on Digital
  libraries}, pages 89--98, 1998.

\bibitem{48}
Prithviraj Sen, Galileo Namata, Mustafa Bilgic, Lise Getoor, Brian Galligher,
  and Tina Eliassi-Rad.
\newblock Collective classification in network data.
\newblock {\em AI magazine}, 29(3):93--93, 2008.

\bibitem{49}
Oleksandr Shchur, Maximilian Mumme, Aleksandar Bojchevski, and Stephan
  G{\"u}nnemann.
\newblock Pitfalls of graph neural network evaluation.
\newblock {\em arXiv preprint arXiv:1811.05868}, 2018.

\bibitem{50}
Nils Kriege and Petra Mutzel.
\newblock Subgraph matching kernels for attributed graphs.
\newblock {\em arXiv preprint arXiv:1206.6483}, 2012.

\bibitem{51}
Nikil Wale, Ian~A Watson, and George Karypis.
\newblock Comparison of descriptor spaces for chemical compound retrieval and
  classification.
\newblock {\em Knowledge and Information Systems}, 14:347--375, 2008.

\bibitem{52}
Aasa Feragen, Niklas Kasenburg, Jens Petersen, Marleen de~Bruijne, and Karsten
  Borgwardt.
\newblock Scalable kernels for graphs with continuous attributes.
\newblock {\em Advances in neural information processing systems}, 26, 2013.

\bibitem{53}
Karsten~M Borgwardt, Cheng~Soon Ong, Stefan Sch{\"o}nauer, SVN Vishwanathan,
  Alex~J Smola, and Hans-Peter Kriegel.
\newblock Protein function prediction via graph kernels.
\newblock {\em Bioinformatics}, 21(suppl\_1):i47--i56, 2005.

\bibitem{54}
Muhan Zhang, Zhicheng Cui, Marion Neumann, and Yixin Chen.
\newblock An end-to-end deep learning architecture for graph classification.
\newblock In {\em Proceedings of the AAAI conference on artificial
  intelligence}, volume~32, 2018.

\bibitem{55}
Andreas Heger, Caleb Webber, Martin Goodson, Chris~P Ponting, and Gerton
  Lunter.
\newblock Gat: a simulation framework for testing the association of genomic
  intervals.
\newblock {\em Bioinformatics}, 29(16):2046--2048, 2013.

\bibitem{56}
Shaked Brody, Uri Alon, and Eran Yahav.
\newblock How attentive are graph attention networks?
\newblock {\em arXiv preprint arXiv:2105.14491}, 2021.

\bibitem{57}
Keyulu Xu, Weihua Hu, Jure Leskovec, and Stefanie Jegelka.
\newblock How powerful are graph neural networks?
\newblock {\em arXiv preprint arXiv:1810.00826}, 2018.

\bibitem{59}
Hao Zhu and Piotr Koniusz.
\newblock Simple spectral graph convolution.
\newblock In {\em International conference on learning representations}, 2021.

\bibitem{wang2024graph}
Zixiao Wang and Jicong Fan.
\newblock Graph classification via reference distribution learning: theory and
  practice.
\newblock {\em Advances in Neural Information Processing Systems},
  37:137698--137740, 2024.

\bibitem{abbahaddougraph}
Yassine Abbahaddou, Fragkiskos~D Malliaros, Johannes~F Lutzeyer, Amine~M
  Aboussalah, and Michalis Vazirgiannis.
\newblock Graph neural network generalization with gaussian mixture model based
  augmentation.
\newblock In {\em Forty-second International Conference on Machine Learning}.

\bibitem{xiong2022pseudo}
Bo~Xiong, Shichao Zhu, Nico Potyka, Shirui Pan, Chuan Zhou, and Steffen Staab.
\newblock Pseudo-riemannian graph convolutional networks.
\newblock {\em Advances in Neural Information Processing Systems},
  35:3488--3501, 2022.

\bibitem{luo2024classic}
Yuankai Luo, Lei Shi, and Xiao-Ming Wu.
\newblock Classic gnns are strong baselines: Reassessing gnns for node
  classification.
\newblock {\em Advances in Neural Information Processing Systems},
  37:97650--97669, 2024.

\bibitem{61}
Li~Sun, Junda Ye, Hao Peng, Feiyang Wang, and Philip~S Yu.
\newblock Self-supervised continual graph learning in adaptive riemannian
  spaces.
\newblock In {\em Proceedings of the AAAI conference on artificial
  intelligence}, volume~37, pages 4633--4642, 2023.

\end{thebibliography}

\end{sloppypar}
\end{document}